\documentclass[preprint,12pt,sort&compress]{elsarticle}

\usepackage{amssymb}
\usepackage{amsmath}
\usepackage[capitalize]{cleveref}

\usepackage[utf8]{inputenc}
\usepackage{newunicodechar}  

\newunicodechar{√}{\ding{51}}  
\usepackage{pifont}  
\journal{Image and Vision Computing}

\begin{document}
\begin{frontmatter}


\begin{highlights}
\item Research highlight 1

A HSI classification method combining multi-scale convolution and Mamba is proposed.
\item Research highlight 2

MCSE module composed of multi-scale convolution and SENet is designed.
\item Research highlight 3

Mamba's long sequence modeling capability is utilized. 
\item Research highlight 4

The residual connection is added before Mamba to prevent original information loss.
\item Research highlight 5

A dual-branch module is proposed to aggregate the central and global information.
\end{highlights}

\title{MSCM-net: A hyperspectral image classification method based on multi-scale convolution and Mamba} 
\author[th,ntu]{Jianjun Chen} 
\ead{chenjianjun@qut.edu.cn}
\affiliation[th]{organization={Qingdao University of Technology
},
            city={Qingdao},
            postcode={266520}, 
            country={China}}
\affiliation[ntu]{organization={Nanyang Technological University},
            city={Singapore},
            postcode={639798},
            country={Singapore}}
\author[th]{Linlin Wang\corref{cor1}}
\ead{wanglinlin@stu.qut.edu.cn}
\cortext[cor1]{Corresponding author}
\author[th]{Lifang Chang}
\author[th]{Limin Huo}
\author[th]{Shujiang Song}
\author[th]{Yanjia Zhao}
\author[th]{Mingwei Shao}
\begin{abstract}
Hyperspectral imaging is widely used in remote sensing and engineering. Therefore, research on its classification methods is crucial. While CNN and Transformer-based methods have advanced, they still face locality constraints and high computational complexity. To address these issues, we propose an innovative hyperspectral image classification model, MSCM-net. Specifically, first of all, a model architecture combining multi-scale CNN and Mamba is proposed. It consists of a multi-scale feature extraction module (MCSE) and multiple stacked Mamba blocks, which integrates the local feature extraction capability of multi-scale CNN and the long sequence modeling advantage of Mamba. Secondly, the proposed MCSE module consists of multi-scale convolution and SENet. Convolution kernels of different scales extract local information with different receptive fields, enhancing the fusion of spatial and spectral information. Meanwhile, the SENet enables the model to automatically learn the importance of each channel in the multi-scale features. Furthermore, we also propose a dual-branch feature aggregation module, which further effectively extracts and integrates the spectral information contained in the central pixel and the spatial information in the surrounding pixels. Our model has undergone numerous experiments on three widely used benchmark datasets. The experimental results show that MSCM-net can achieve advanced classification performance while reducing computational complexity.
\end{abstract}


\end{frontmatter}




\section{Introduction}
By combining imaging techniques with spectroscopy, hyperspectral imaging (HSI) enables the concurrent capture of a target's spatial distribution and its spectral signature \cite{GaoJingpeng2026,ZhaoJinling2025,WangXinya2023}. The inherent fusion of spatial-spectral characteristics within hyperspectral datasets provides superior discriminative power for remote sensing classification tasks, particularly in domains such as agriculture \cite{Yang2024}, environmental and climate monitoring \cite{Ali2024}, urban development \cite{WangJinfeng2023} and military security \cite{Song2023}. However, although the spectral diversity enhances the classification potential, due to the high dimensionality of the data and the need to effectively extract spectral and spatial context information, HSI classification remains a challenging issue.

In recent years, various machine learning and deep learning techniques have been proposed to meet these challenges. Traditional machine learning models, such as Support Vector Machine (SVM) \cite{DuPeijun2012} and Random Forest (RF) \cite{ZhangYouqiang2018} have long been applied to HSI classification. For instance, in 2022, Liu et al.\cite{LiuGuangxin2022} proposed a non-parallel support vector machine model and applied it to hyperspectral image classification. However, these methods have limited effectiveness. The main reason is that they couldn't fully utilize the spectral space characteristics of the data. In contrast, CNN that has emerged in recent years \cite{ZhouDing-Xuan2018,YasinMagombe2024,AthaDeegan2018} utilizes hierarchical convolution to extract spatial features and has shown promising results, outperforming traditional methods significantly. For instance, Yang et al.\cite{YangXiaofei2018} proposed four new deep learning models, namely the two-dimensional convolutional neural network (2-D-CNN) the three-dimensional CNN, the recurrent two-dimensional CNN (R-2-D-CNN), and the recurrent three-dimensional CNN (R-3-D-CNN) for hyperspectral image classification. Paul et al.\cite{PaulArati2021} proposed a novel deep learning framework, which efficiently utilized CNN and spatial pyramid pooling (SPP) to extract spectral-spatial features for classification. The results obtained demonstrated the superiority of the proposed model in effectively classifying hyperspectral images (HSI). Feng et al.\cite{FengFan2019} designed an 11-layer CNN model, named R-HybridSN (Residual-Hybrid SN). This model organically integrated 3D-2D-CNN, residual learning and depthwise separable convolution, and was capable of better learning deep spatial-spectral features with extremely limited training data. Lei et al.\cite{LeiXiaohan2025} proposed a graph CNN based on the angular field (GAF-SGCN) in 2025 for hyperspectral image classification, which could more effectively extract and fuse the deep features in the image. However, the above-mentioned CNN-based methods are all restricted by the local receptive field, which to some extent limits their ability to capture long-range dependencies in hyperspectral data.

To address this issue, researchers have incorporated Transformer model into HSI. Transformer model utilizes the self-attention mechanism to capture distant dependencies, and by enabling the modeling of global relationships across input data, has achieved considerable success in the HSI field. For instance, Hu et al.\cite{HuXiang2021} proposed an unsupervised hyperspectral image classification framework that combined the contrast learning method with Transformer. This model could effectively extract the features of hyperspectral images in an unsupervised manner. Tan et al.\cite{TanYunfei2024} proposed a hyperspectral image classification method based on the Embedded Linear Vision Transformer (ELViT), which significantly improved the classification performance of the model. Fu et al.\cite{FuChuan2025} proposed a small-sample hyperspectral classification algorithm based on CTA-net in 2025. By combining CNN and Transformer, they alleviated the problem of insufficient samples. Fan et al.\cite{FanXiao-yong2025} proposed a fine classification method for rare earth mining areas based on object-oriented thinking and multi-layer attention convolutional neural network (OCTC) in 2025. They integrated Transformer and CRAM to enhance the reliability of the model, as well as the feature extraction ability and overall classification accuracy. However, Transformer has a significant drawback: the computational complexity of its self-attention mechanism increases quadratically with the size of the input. This high computational cost limits the practicality of models based on Transformer, especially for large-scale hyperspectral datasets that require real-time or resource-efficient processing.

To address these computational challenges, the state space model (SSM) \cite{Schuessler2022} , as the latest proposed model capable of long sequence modeling, has attracted attention. SSM is commonly used in the analysis of dynamic systems and the modeling of time series. It is adept at mapping sequential data to the state space to capture long-term dependencies. Mamba \cite{SunJunding2025} represents the latest advancement in architecture-based SSM, introducing the selective state space model that achieves more efficient long sequence modeling by selectively retaining or discarding information based on the data context. This innovation has demonstrated state-of-the-art performance in areas such as language modeling, audio processing \cite{ZhangChongbin2025}, and image understanding. For instance, Wang et al.\cite{WangHao2025} proposed a dual-branch hybrid architecture named DBMGNet in 2025, which combined Mamba and GCN for hyperspectral image classification. Jiang et al.\cite{JiangYonghua2025} proposed a structurally enhanced spatial-spectral dynamic gating Mamba (SEDGM) method in 2025, which utilized the collaborative design of the spatial and spectral gating Mamba mechanisms to extract and utilize the key regional features of hyperspectral data. Li et al.\cite{LiYapeng2024} proposed the first image-level HSI classification model based on Mamba in 2025. They extracted HSI image features through a spatial-spectral dual-branch approach. Liang et al.\cite{LiangLianhui2025} proposed a new dual-branch class Mamba linear attention (DBMLLA) network in 2025, which achieved efficient modeling of global dependencies.

In conclusion, compared to Transformers, CNNs offer higher computational efficiency and superior local feature extraction. This study combines Mamba with CNN to leverage Mamba's long-sequence modeling capability while enhancing local feature extraction without increasing computational complexity. We propose MSCM-net, a hyperspectral image classification framework integrating multi-scale convolution and Mamba. First, multi-scale convolution with different kernel sizes performs multi-branch feature extraction to enhance local information focus. Then, stacked Mamba blocks construct a long-sequence model for deeper feature extraction. Experiments demonstrate that this simple combination achieves satisfactory results in hyperspectral image classification.

The main contributions of this work can be summarized as follows:

(1) We propose a HSI classification architecture based on multi-scale convolution and Mamba (MSCM-net). It utilizes Mamba's sequence modeling capability to establish the dependency relationship between the central pixel and the large number of neighboring pixels. Furthermore, the multi-scale convolution enhances the ability to extract spatial local information. Compared with traditional attention models such as Transformer, the computational complexity of the MSCM-net structure is reduced, thereby improving the computational efficiency. 

(2) The SENet is added to the MCSE module. By learning the importance weights of different scale feature extraction channels, adaptive feature re-calibration is achieved, thereby further enhancing the effect of multi-scale feature fusion.

(3) To further enhance the extraction of spectral and spatial information, we have also proposed a dual-branch feature aggregation module. It achieves in-depth exploration and fusion of the features of the hyperspectral image through two parallel feature extraction paths - central feature extraction and global feature extraction. Two paths respectively extract features from the local details and global information dimensions. Feature fusion is achieved through simple element-wise addition, resulting in a final feature representation that is both discriminative and robust.

(4) A large number of experiments conducted on three benchmark hyperspectral datasets (Indian Pines, WHU-Hi-HongHu and Salinas) have demonstrated the superiority of MSCM-net. A large number of ablation experiments have also verified the effectiveness of the framework components, fully demonstrating that MSCM-net can achieve advanced classification results while significantly reducing computational costs.

\section{METHODOLOGY}
\subsection{Model architecture}
\begin{figure}[htbp]
    \centering
    \includegraphics[width=1\linewidth]{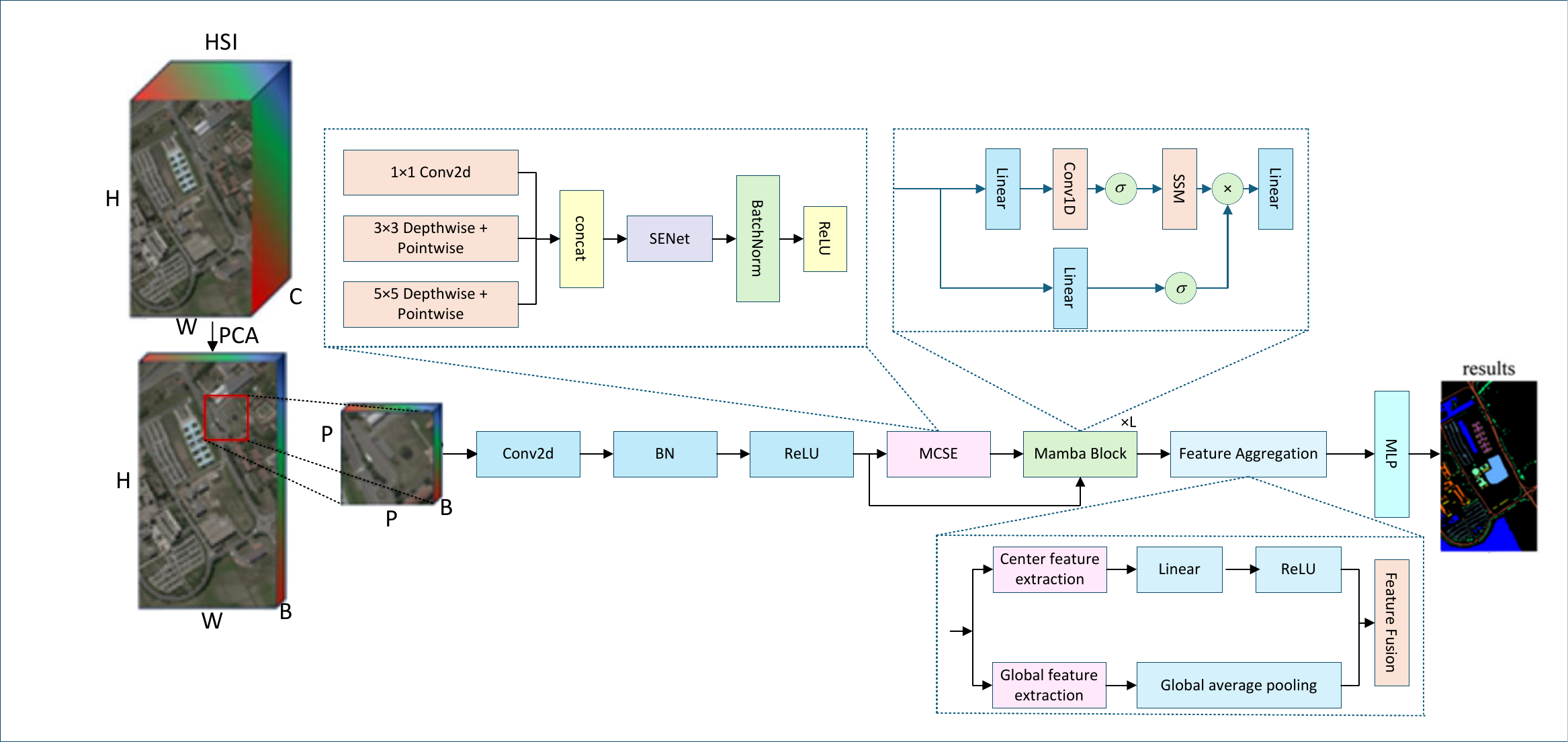}  
    \caption{Overall architecture of the MSCM-net model}
    \label{figure1}
\end{figure}

The overall framework of the model proposed in this paper is shown in Figure \ref{figure1}, which mainly consists of three parts: the multi-scale convolutional feature extraction module (MCSE), the Mamba feature extraction module, and the dual-branch feature aggregation module.

Given a hyperspectral image, denoted as $H\in {{R}^{B\times H\times W}}$, where $H$ and $W$ represent the height and width of the image, respectively, and $B$ is the number of spectral channels. In the preprocessing stage, principal component analysis (PCA) is applied to the HSI data to obtain $X\in {{R}^{C\times H\times W}}$, where $C$ represents the number of spectral dimensions after PCA. After applying PCA, each pixel is taken as the center, and patches of the given size P (in this case, 15) are divided. For the edge pixels, zero padding is applied to maintain the patch size. Therefore, each sample is $Z\in {{R}^{C\times P\times P}}$, and the goal is to use these patches to classify the corresponding central pixels.

After obtaining the training and testing samples, the first stage of the model is spectral feature compression \cite{KongFanqiang2024}, which is a crucial preprocessing step for the model. Through learning and weighted combination, the original multiple redundant bands are compressed into 64 most discriminative feature channels, eliminating the correlations and redundancies among the spectra. Using a $3\times 3$ convolution kernel, local spatial context information is initially captured to ensure that the subsequent stages can focus on more complex feature representations.

After obtaining 64 basic feature channels, the model performs multi-scale spatial feature extraction through the MCSE module. The MCSE module adopts a three-branch parallel strategy, namely $1\times 1$ convolution, $3\times 3$ depthwise separable convolution, and $5\times 5$ depthwise separable convolution. The $1\times 1$ convolution focuses on the global interaction between channels, the $3\times 3$ depthwise separable convolution captures local texture details, and the $5\times 5$ depthwise separable convolution extracts larger spatial context information. The outputs of the three branches are concatenated along the channel dimension, resulting in 192 feature channels. Subsequently, the SENet performs adaptive weighting on the features of these three branches. By learning the importance weights of each channel, it highlights the channels that are crucial for the classification task and suppresses irrelevant features. This multi-scale design enables the model to simultaneously perceive spatial and spectral features at different scales, while the attention mechanism ensures the efficiency of feature utilization.

The enhanced features need to be converted into a format suitable for sequence processing and then input into the Mamba block for further feature extraction. The model reorganizes the spatial feature maps into a sequential form, expanding the $13\times 13$ grid into a sequence of 169 positions, with each position corresponding to a 128-dimensional feature vector. Meanwhile, the 64-dimensional features after the original features compressed are also projected and adjusted to the same dimension. The characteristics of the two paths are fused through the residual connection and additional learnable positional encoding is added to form the input of the Mamba module. Four stacked Mamba encoder blocks gradually deepen the feature representation. Each block contains residual connections, layer normalization, and SiLU activation, ensuring stable training and the effective transmission of feature information.

After deep Mamba modeling, the model employs a dual-path feature aggregation strategy combining central and global features. Based on the prior knowledge that central pixels contain the most comprehensive category information in hyperspectral classification, the central position features are emphasized. Simultaneously, global average pooling over the entire sequence captures overall image statistics. The fused features are fed into a classification multi-layer perceptron, where batch normalization, Dropout regularization, and ReLU activation are applied before projection to the output space corresponding to target categories, generating prediction scores for final hyperspectral image classification.
\subsection{MCSE Module}
The MCSE module is one of the core architectures of the entire model. MCSE receives the output features from the spectral feature compression layer${{X}_{original}}\in {{R}^{B\times 64\times W\times H}}$, where $B$ represents the batch size, $W=H=13$ is the spatial dimension, and 64 represents the number of feature channels. The module performs feature extraction through three independent branches. The first branch uses $1\times 1$ convolution kernels for pure channel mixing, without changing the spatial dimensions but recombining the spectral feature channels:
\begin{equation}
{{X}_{1}}={{W}_{1\times 1}}*{{X}_{original}}+{{b}_{1\times 1}}
\end{equation}
Where ${{W}_{1\times 1}}\in {{R}^{N\times 64\times 1\times 1}}$ represents the convolution weights, ${{b}_{1\times 1}}$ represents the bias term, and $N=64$ represents the number of output channels. The second branch employs a $3\times 3$ depthwise separable convolution structure to extract local details. Firstly, group convolution is used to extract local spatial features, and then $1\times 1$ convolution is applied to achieve channel interaction:
\begin{equation}
X_{2,depth}^{(c)}=W_{dw3}^{(c)}*X_{original}^{(c)},\quad c=1,...,64
\end{equation}
\begin{equation}
{{X}_{2}}=\operatorname{Re}LU(B{{N}_{3\times 3}}({{W}_{pw3}}*{{X}_{2,depth}}+{{b}_{pw3}}))
\end{equation}
Where $W_{dw3}^{(c)}\in {{R}^{1\times 1\times 3\times 3}}$ represents the deep convolution kernel, and ${{W}_{pw3}}\in {{R}^{N\times 64\times 1\times 1}}$ represents the point-wise convolution weights. The third branch is structurally similar to the second branch, but uses a larger $5\times 5$ convolution kernel to capture broader contextual information:
\begin{equation}
X_{3,depth}^{(c)}=W_{dw5}^{(c)}*X_{original}^{(c)},\quad c=1,...,64
\end{equation}
\begin{equation}
{{X}_{3}}=\operatorname{Re}LU(B{{N}_{5\times 5}}({{W}_{pw5}}*{{X}_{3,depth}}+{{b}_{pw5}}))
\end{equation}
Where $W_{dw5}^{(c)}\in {{R}^{1\times 1\times 5\times 5}}$ is a $5\times 5$ deep convolution kernel.

The features extracted by the three branches are concatenated along the channel dimension to form multi-scale fused features:
\begin{equation}
{{X}_{cat}}=concat({{X}_{1}},{{X}_{2}},{{X}_{3}})\in {{R}^{B\times 3N\times W\times H}}
\end{equation}
Where $3N=192$ represents the total number of channels after concatenation.

Furthermore, to further enhance the discriminative ability of the features, the MCSE module also introduces the Squeeze-and-Excitation channel attention mechanism (SENet) to adaptively weight the concatenated features. The attention mechanism consists of two stages. The first stage is the compression stage, where the global average pooling is used to obtain the global statistical features of each channel:
\begin{equation}
{{Z}_{c}}=\frac{1}{W\times H}\sum\limits_{i=1}^{W}{\sum\limits_{j=1}^{H}{X_{cat}^{(c)}}}(i,j),\quad Z\in {{R}^{B\times 3N\times 1\times 1}}
\end{equation}

The second stage is the incentive stage, where the importance weights of each channel are learned through two fully connected layers:
\begin{equation}
S=\sigma ({{W}_{2}}\centerdot \operatorname{Re}LU({{W}_{1}}\centerdot Z+{{b}_{1}})+{{b}_{2}})
\end{equation}

Where $S\in {{R}^{B\times 3N\times 1\times 1}}$ represents the attention weights for each channel. $\sigma $ is the Sigmoid activation function. ${{W}_{1}}$ and ${{W}_{2}}$ are the weights of the fully connected layer. Then, the learned attention weights are multiplied with the original features channel by channel to highlight important features and suppress irrelevant ones:
\begin{equation}
{{X}_{att}}={{X}_{cat}}\odot S
\end{equation}

Where $\odot $ represents the element-wise multiplication operation, and the attention weight $S$ is expanded to the $W$ spatial dimension through the broadcast mechanism. Finally, the weighted features are subjected to batch normalization and ReLU activation for output. The output of the MCSE module will serve as the input for subsequent serialization processing and Mamba modeling. It is combined with the original Conv2d features through residual connections to form complementary feature representations, providing rich spatial context information for hyperspectral image classification.
\subsection{Mamba}
After extracting multi-scale spatial features in the MCSE module, the model converts the spatial feature maps into a sequence form to adapt to the sequence processing of the Mamba module. This process involves the feature processing and fusion of two parallel paths. First, the spatial feature maps of the MCSE module are converted into a sequence form, as shown in Equation \ref{10} :
\begin{equation}
{{X}_{1\_seq}}=rearrange({{X}_{MCSE\_out}})\in {{R}^{B\times L\times \dim}}
\label{10}
\end{equation}
Where $L=169$ represents the sequence length. $\dim=128$ represents the feature dimension, and the "rearrange" operation expands the $13\times 13$ spatial grid into 169 position sequences. The original features extracted by the initial Conv2d are also processed in a sequential manner:
\begin{equation}
{{X}_{orig\_seq}}=rearrange({{X}_{original}})\in {{R}^{B\times L\times 64}}
\end{equation}
\begin{equation}
{{X}_{orig\_proj}}=\operatorname{Re}LU(Dropout({{W}_{proj}}{{X}_{orig\_seq}}+{{b}_{proj}}))\in {{R}^{B\times L\times \dim}}
\end{equation}
Where ${{W}_{proj}}\in {{R}^{\dim\times 64}}$ represents the projection matrix, and ${{b}_{proj}}$ is the bias term.

The characteristics of the two paths are fused through residual connections, and are further augmented with learnable positional encodings:
\begin{equation}
{{X}_{fused}}=Dropout({{X}_{orig\_proj}}\oplus {{X}_{MCSE\_seq}})
\end{equation}
\begin{equation}
{{X}_{mamba\_in}}={{X}_{fused}}\oplus ({{E}_{pos}}\centerdot \alpha )\in {{R}^{B\times L\times \dim}}
\end{equation}

Where $\oplus $ represents element-wise addition, ${{E}_{pos}}$ is a learnable absolute position encoding matrix, and $\alpha $ is a scaling factor that controls the intensity of position information.

This paper employs stacked Mamba encoder blocks for deep sequence modeling. Each Mamba block has a structure that includes a selective state space layer, layer normalization, SiLU activation, and residual connections. Among them, Mamba is based on the selective state space model, and its discretized state space equation can be expressed as:
\begin{equation}
{{h}_{t}}={{\bar{A}}_{t}}{{h}_{t-1}}+{{\bar{B}}_{t}}{{x}_{t}}
\end{equation}
\begin{equation}
{{y}_{t}}={{C}_{t}}{{h}_{t}}+{{D}_{t}}{{x}_{t}}
\end{equation}

Where ${{h}_{t}}$ represents the hidden state at time t, ${{x}_{t}}$ is the input,${{y}_{t}}$ is the output and ${{D}_{t}}$ is the state dimension. Unlike traditional SSM, the parameters of Mamba $\Delta t,A,{{B}_{t}},{{C}_{t}}$ are all functions of the input${{x}_{t}}$, implementing a selective scanning mechanism:
\begin{equation}
\Delta t=softplus({{W}_{\Delta }}{{x}_{t}}+{{b}_{\Delta }})
\end{equation}
\begin{equation}
{{B}_{t}}=linea{{r}_{B}}({{x}_{t}})
\end{equation}
\begin{equation}
{{C}_{t}}=linea{{r}_{C}}({{x}_{t}})
\end{equation}

Where ${{W}_{\Delta }}$ represents the weight of the time-step parameterization matrix. This enables the parameterization that depends on the input to allow the model to dynamically adjust the state transition based on the current input, thereby enhancing its ability to focus on important information in long sequences.

Continuous-time parameters are discretized using the zero-order hold (ZOH) method:
\begin{equation}
{{\bar{A}}_{t}}=\exp ({{\Delta }_{t}}A)
\end{equation}
\begin{equation}
{{\bar{B}}_{t}}={{\Delta }_{t}}{{B}_{t}}(\exp ({{\Delta }_{t}}A)-I)\centerdot {{A}^{-1}}
\end{equation}

The output of the Mamba module serves as the basis for subsequent feature extraction and classification. Its rich serialized feature representation provides a powerful modeling capability for the hyperspectral image classification task.
\subsection{Dual-branch feature aggregation module}
This paper adopts a central-global dual-branch feature aggregation method, where the central feature extraction and global feature extraction branches are used to further extract the features obtained from the MCSE and Mamba modules. The central pixel of the image block usually contains the most comprehensive category information. Therefore, a center feature extraction module is designed specifically to extract the feature representation of the central position from the Mamba output sequence. First, determine the central position. For a sequence of length $L$, the index of the central position is calculated as follows:
\begin{equation}
{{c}_{idx}}=\frac{L}{2}
\end{equation}

This position corresponds to the central pixel of the original $13\times 13$ spatial grid.
Extract the feature vector at the ${{c}_{idx}}$th position from the Mamba output sequence:
\begin{equation}
{{X}_{center}}={{X}_{mamba\_out}}[:,{{c}_{idx}},:]\in {{R}^{B\times \dim}}
\end{equation}

Where ${{X}_{mamba\_out}}$ represents the output of the Mamba block, and ${{X}_{center}}$ is the extracted central feature matrix, which contains the central pixel features of all samples in the batch. Then, a small feature enhancement network is used to further refine the central features:

\begin{equation}
{{X}_{center\_feat}}=\operatorname{Re}LU(Dropout({{W}_{center}}{{X}_{center}}+{{b}_{center}}))\in {{R}^{B\times \dim}}
\end{equation}
Where ${{W}_{center}}\in {{R}^{\dim\times \dim}}$ represents the weight matrix and ${{b}_{center}}$ represents the bias term. This enhancement layer further exploits the discriminative information of the central features through a nonlinear transformation.

Another branch is the global feature extraction path, which obtains statistical feature information from the entire image block. The specific steps are to perform an average operation along the sequence dimension of the Mamba output sequence to obtain the global statistical features of the entire image block:
\begin{equation}
{{X}_{global}}=\frac{1}{L}\sum\limits_{t=1}^{L}{{{X}_{mamba\_out}}}[:,t,:]\in {{R}^{B\times \dim}}
\end{equation}

The features obtained through center feature extraction and global feature extraction are directly added together after passing through the feature fusion module:
\begin{equation}
{{X}_{com}}={{X}_{center\_feat}}\oplus {{X}_{global}}\in {{R}^{B\times \dim}}
\end{equation}
Where $\oplus $ indicates element-wise addition.

This module provides local detail information through the central feature, and global feature information to offer contextual statistical information. The combination of these two achieves information complementarity. The fused feature ${{X}_{com}}$ will be used as the input for the classification head, and is further mapped to the category space through a multi-layer perceptron \cite{WuJiann-Ming2008} to complete the final classification decision. 

\section{Results}

\subsection{Datasets introduction}
(1) Indian Pines

\begin{figure}[htbp]
    \centering
    \includegraphics[width=1\linewidth]{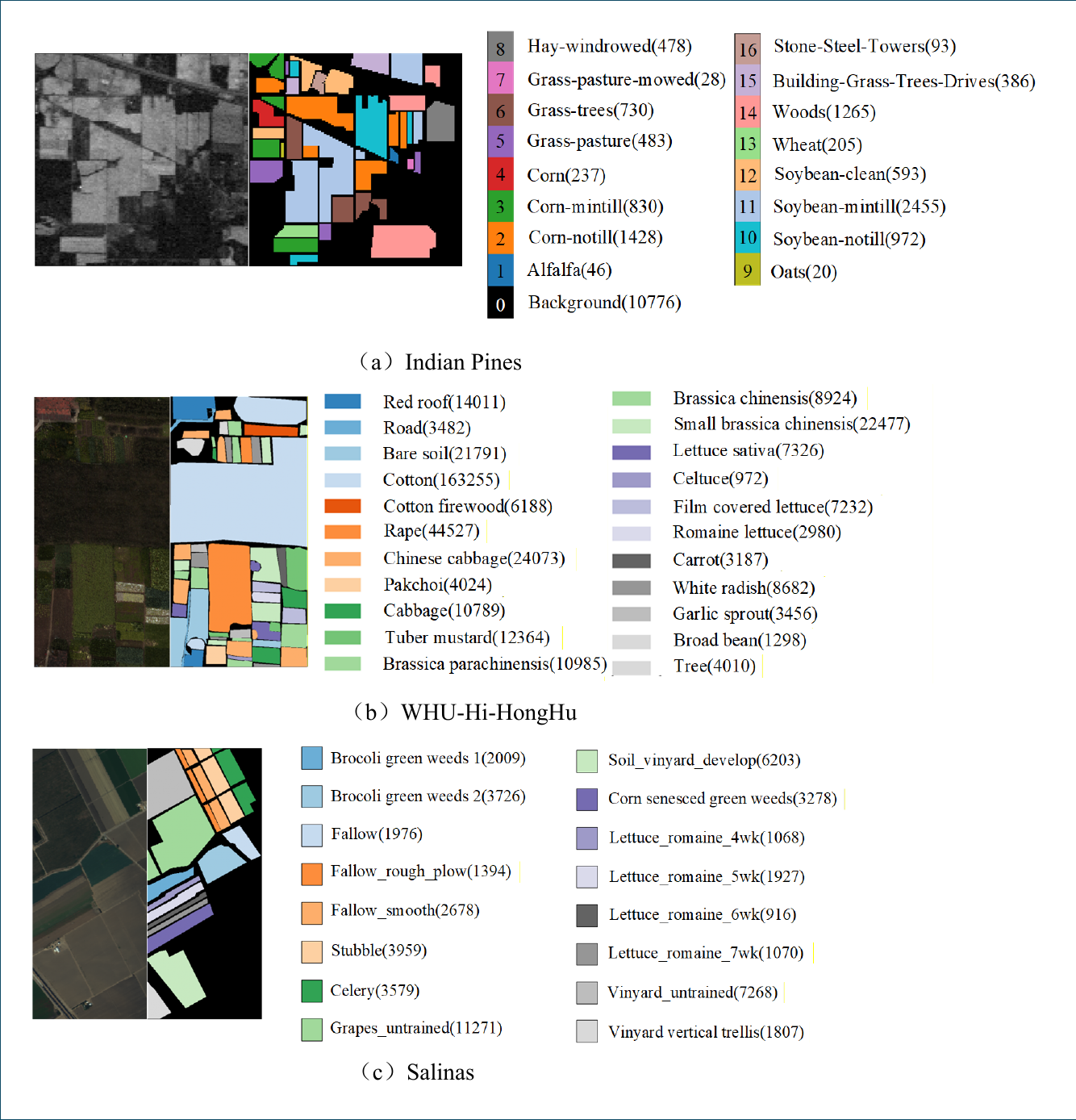}  
    \caption{Images and category information of the three datasets. (a) Indian Pines. (b)WHU-Hi-HongHu. (c)Salinas}
    \label{figure2}
\end{figure}

The Indian Pines dataset was collected using the airborne visible/infrared imaging spectrometer (AVIRIS) sensor over the fields in the northwest of Indiana, USA. This dataset consists of $145\times 145$ pixels and contains 220 spectral bands, with the wavelength range spanning from 0.4 to 2.5 µm. After removing the noise and water absorption bands, 200 bands are retained for classification. The ground is divided into 16 land cover types, which represent different types of crops and vegetation. The images and category information of the Indian Pines dataset are shown in (a) of Figure \ref{figure2}, and the distribution of training and test samples is shown in Table \ref{table1}.

\begin{table}[htbp]
\caption{Training and testing distribution of the Indian Pines dataset, the WHU-Hi-HongHu dataset, and the Salinas dataset.}\label{table1}
\centering
\scriptsize  
\setlength{\tabcolsep}{1.8pt}  
\renewcommand{\arraystretch}{0.9}  
\begin{tabular}{lll|lll|lll}
\hline
\multicolumn{3}{c|}{Indian Pines}                                                    & \multicolumn{3}{c|}{WHU-Hi-HongHu}                                                   & \multicolumn{3}{c}{Salinas}                                                         \\
\multicolumn{1}{c}{Category} & \multicolumn{1}{c}{Train} & \multicolumn{1}{c|}{Test} & \multicolumn{1}{c}{Category} & \multicolumn{1}{c}{Train} & \multicolumn{1}{c|}{Test} & \multicolumn{1}{c}{Category} & \multicolumn{1}{c}{Train} & \multicolumn{1}{c}{Test} \\ \hline
Alfalfa                      & 30                        & 16                        & Red roof                     & 30                        & 13981                     & Brocoli\_green\_weeds\_1     & 30                        & 1979                     \\
Corn-notill                  & 30                        & 1398                      & Road                         & 30                        & 3452                      & Brocoli\_green\_weeds\_2     & 30                        & 3696                     \\
Corn-mintill                 & 30                        & 800                       & Bare soil                    & 30                        & 21761                     & Fallow                       & 30                        & 1946                     \\
Corn                         & 30                        & 207                       & Cotton                       & 30                        & 163225                    & Fallow\_rough\_plow          & 30                        & 1364                     \\
Grass-pasture                & 30                        & 453                       & Cotton   firewood            & 30                        & 6158                      & Fallow\_smooth               & 30                        & 2648                     \\
Grass-trees                  & 30                        & 700                       & Rape                         & 30                        & 44497                     & Stubble                      & 30                        & 3929                     \\
Grass-pasture-mowed          & 14                        & 14                        & Chinese   cabbage            & 30                        & 24043                     & Celery                       & 30                        & 3549                     \\
Hay-windrowed                & 30                        & 448                       & Pakchoi                      & 30                        & 3994                      & Grapes\_untrained            & 30                        & 11241                    \\
Oats                         & 10                        & 10                        & Cabbage                      & 30                        & 10759                     & Soil\_vinyard\_develop       & 30                        & 6173                     \\
Soybean-notill               & 30                        & 942                       & Tuber   mustard              & 30                        & 12334                     & Corn\_senesced\_green\_weeds & 30                        & 3248                     \\
Soybean-mintill              & 30                        & 2425                      & Brassica   parachinensis     & 30                        & 10955                     & Lettuce\_romaine\_4wk        & 30                        & 1038                     \\
Soybean-clean                & 30                        & 563                       & Brassica   chinensis         & 30                        & 8894                      & Lettuce\_romaine\_5wk        & 30                        & 1897                     \\
Wheat                        & 30                        & 175                       & Small   brassica chinensis   & 30                        & 22447                     & Lettuce\_romaine\_6wk        & 30                        & 886                      \\
Woods                        & 30                        & 1235                      & Lactuca   sativa             & 30                        & 7296                      & Lettuce\_romaine\_7wk        & 30                        & 1040                     \\
Buildings-Grass-Trees        & 30                        & 356                       & Celtuce                      & 30                        & 942                       & Vinyard\_untrained           & 30                        & 7238                     \\
Stone-Steel-Towers           & 30                        & 63                        & Film   covered lettuce       & 30                        & 7202                      & Vinyard\_vertical\_trellis   & 30                        & 1777                     \\
                             &                           &                           & Romaine   lettuce            & 30                        & 2950                      &                              &                           &                          \\
\multicolumn{1}{c}{}         & \multicolumn{1}{c}{}      & \multicolumn{1}{c|}{}     & Carrot                       & 30                        & 3157                      & \multicolumn{1}{c}{}         & \multicolumn{1}{c}{}      & \multicolumn{1}{c}{}     \\
\multicolumn{1}{c}{}         & \multicolumn{1}{c}{}      & \multicolumn{1}{c|}{}     & White   radish               & 30                        & 8652                      & \multicolumn{1}{c}{}         & \multicolumn{1}{c}{}      & \multicolumn{1}{c}{}     \\
\multicolumn{1}{c}{}         & \multicolumn{1}{c}{}      & \multicolumn{1}{c|}{}     & Garlic   sprout              & 30                        & 3426                      & \multicolumn{1}{c}{}         & \multicolumn{1}{c}{}      & \multicolumn{1}{c}{}     \\
\multicolumn{1}{c}{}         & \multicolumn{1}{c}{}      & \multicolumn{1}{c|}{}     & Broad   bean                 & 30                        & 1268                      & \multicolumn{1}{c}{}         & \multicolumn{1}{c}{}      & \multicolumn{1}{c}{}     \\
\multicolumn{1}{c}{}         & \multicolumn{1}{c}{}      & \multicolumn{1}{c|}{}     & Tree                         & 30                        & 3980                      & \multicolumn{1}{c}{}         & \multicolumn{1}{c}{}      & \multicolumn{1}{c}{}     \\ \hline
Total                        & 444                       & 9805                      & Total                        & 660                       & 386003                    & Total                        & 480                       & 53649                    \\ \hline
\end{tabular}
\end{table}

(2) WHU-Hi-HongHu

The WHU-Hi-HongHu dataset was collected using the WHU hyperspectral imager. And the target area is an agricultural zone near Honghu City, China. This dataset consists of 940×475 pixels and has 270 spectral bands, capturing complex land cover types such as bare soil, cotton fields, cabbages, and trees, etc. The dataset includes 22 land cover categories, making it the largest and most diverse dataset in this study. The image and category information of the WHU-Hi-HongHu dataset is shown in (b) Figure \ref{figure2}, and the distribution of training and test samples is shown in Table \ref{table1}.

(3) Salinas

The Salinas dataset is an airborne hyperspectral remote sensing dataset collected in the Salinas Valley area of California, USA, using the AVIRIS sensor. It consists of 224 spectral bands, with a wavelength range of 400 to 2500 nm. The image size is 512×217 pixels and the spatial resolution is 3.7 m/pixel. The dataset contains 54,129 labeled pixels with 16 types of land cover (such as Fallow, Celery, etc.). At the same time, the distribution of the number of local land cover types in this dataset is relatively balanced, and the background is relatively simple. The image and category information of the Salinas dataset are shown in (c) of Figure \ref{figure2}, and the distribution of training and test samples is shown in Table \ref{table1}.

\subsection{Experimental setup}
In this study, the experiments are implemented and trained using the PyTorch framework. During the training process, the NVIDIA GeForce GTX 4090 GPU (with 24GB of memory) is used for acceleration. The relevant settings for implementation are as follows: an Adam optimizer with a learning rate of 1e-3, a training batch size of 64 for the model, a total of 100 epochs, a Mamba block depth of 4, and a patch size of 15.

To verify the effectiveness of the proposed method, this paper evaluates it by using many different HSI classification methods as a comparison. Including methods based on machine learning, methods based on CNN, methods based on Transformer, and methods based on Mamba, the introduction of the comparison methods is as follows:

(1) SVM \cite{TanKun2008}: This method solved the problem of linearly inseparable data through kernel function mapping, achieving pixel-level land classification.

(2) 3DCNN \cite{FiratHuseyin2023}: This method employed a 3DCNN-based ResNet50 combined with PCA for dimensionality reduction.

(3) SSRN \cite{ZhongZilong2018}: The spectral-spatial Residual Network (SSRN) took a three-dimensional cube as input and used spectral and spatial residual blocks for identity mapping and batch normalization to improve the classification accuracy of hyperspectral images. It was an end-to-end network.

(4) SSFTT \cite{SunLe2022}: The Spectral-Spatial Feature Transformer (SSFTT) integrated 3D and 2D convolutional layers for low-level feature extraction. Gaussian weighted markers were used for feature transformation, and transformer encoders were employed to capture deep spectral space and semantic features for hyperspectral image classification.

(5) MASSFormer \cite{SunLe2024}: Memory-enhanced Spectral-Space Transformer (MASSFormer) combined a memory marker to store previous spectral-space features and a memory-enhanced transformer encoder to enhance information mixing and improve classification accuracy.

(6) SQSFormer \cite{ChenNing2024}: This method, while preserving the spectral features, utilized the adaptive query of the central pixel to access relevant spatial information. It combined rotation-invariant positional embedding and the center-focused module to enhance the integration of spectral and spatial information, thereby improving the classification accuracy of hyperspectral images.

(7) SSMamba \cite{HuangLingbo2024}: This method effectively modeled the remote spectral spatial dependencies by using spectral space tokens generation and stacking Mamba blocks, thereby enhancing the classification ability of hyperspectral data. It achieved faster processing speed and lower computational complexity compared to Transformer.
\subsection{Experimental results}

The classification results of the three datasets used in this article by different methods reported in the literature are shown in Tables \ref{table2}-\ref{table4}. As can be seen from the tables, when the training samples are set to 30, the method proposed in this article achieves the best classification effect in all three datasets.
\begin{figure}[htbp]
    \centering
    \includegraphics[width=1\linewidth]{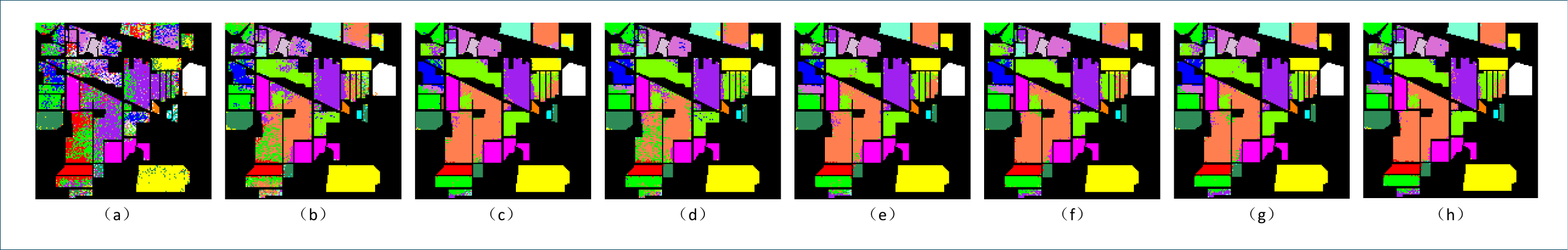}  
    \caption{Classification results of different methods on the Indian Pines dataset (a)SVM (b)3DCNN (c)SSRN (d)SSFTT (e)MASSFormer (f)SQSFormer (g)SSMamba (h) ours}
    \label{figure3}
\end{figure}
\begin{table}[htbp]
\centering
\footnotesize  
\setlength{\tabcolsep}{3pt}  
\caption{Test results (averages) of data classification on the Indian Pines dataset}\label{table2}
\begin{tabular}{@{}l*{8}{c}@{}}
\hline
\multicolumn{1}{c}{Class}  & SVM   & 3DCNN & SSRN  & SSFTT & MASSFormer & SQSFormer & SSMamba & Ours  \\ \hline
1      & 98.75 & 100   & 100   & 100   & 100        & 98.75     & 100     & \textbf{100}   \\
2      & 58.54 & 86.35 & 87.18 & \textbf{90.21}  & 86.98      & 84.18     & 83.46   & 90.20 \\
3      & 67.05 & 81.50 & 87.70 & 91.60 & 88.92      & \textbf{93.03}      & 91.67   & 92.13 \\
4      & 90.05 & \textbf{99.76} & 99.61 & 99.13 & 99.52      & 96.33     & 99.42   & 99.03 \\
5      & 67.86 & 91.39 & 94.75 & 91.08 & 92.36      & 93.16     & 94.57   & \textbf{96.25} \\
6      & 98.49 & 96.76 & 98.00 & 97.06 & 98.63      & 99.54     & 99.43   & \textbf{99.57} \\
7      & 75.05 & 100   & 100   & 100   & 100        & 100       & 98.46   & \textbf{100}   \\
8      & 98.39 & 100   & 99.82 & 99.73 & 97.95      & 100       & 100     & \textbf{100}    \\
9      & 78.72 & 100   & 100   & 100   & 100        & 98.00     & 100     & \textbf{100}   \\
10     & 60.83 & 87.90 & 90.51 & 92.27 & 91.40      & \textbf{92.95}     & 88.68   & 85.03 \\
11     & 66.19 & 83.15 & 86.32 & 79.69 & 87.70      & 90.01     & \textbf{90.99}   & 90.60 \\
12     & 40.00 & 89.28 & 94.07 & 89.73 & 92.75      & \textbf{96.23}     & 94.74   & 90.59 \\
13     & 98.74 & 97.14 & 98.74 & 98.63 & 98.86      & 99.54     & 100     & \textbf{100}   \\
14     & 72.57 & 95.52 & 97.20 & 94.90 & 97.96      & 98.62     & 99.37   & \textbf{99.51} \\
15     & 75.67 & 97.66 & 98.71 & 99.16 & 98.71      & 99.61     & 99.38   & \textbf{99.72} \\
16     & 99.68 & 98.68 & 99.05 & 98.41 & 97.78      & \textbf{100}       & 99.05   & 98.41 \\ \hline
OA(\%) & 69.34 & 89.26 & 91.70 & 90.18 & 92.06      & 93.19     & 91.59   & \textbf{94.35} \\
AA(\%) & 68.30 & 94.07 & 95.73 & 82.60 & 95.60      & 96.25     & 95.46   & \textbf{96.31} \\ 
K×100  & 65.36 & 87.79 & 90.56 & 88.84 & 90.97      & 92.22     & 90.42   & \textbf{93.03} \\ \hline
\end{tabular}
\end{table}
\begin{figure}[htbp]
    \centering
    \includegraphics[width=1\linewidth]{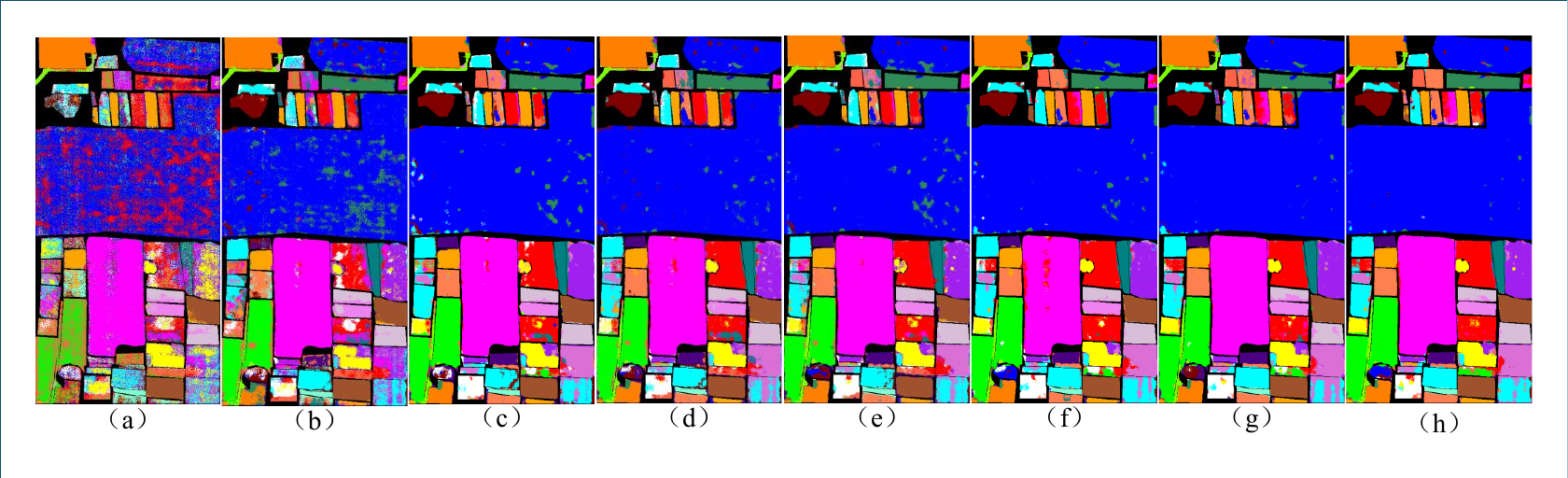}  
    \caption{Classification results using different methods on the WHU-Hi-HongHu dataset (a)SVM (b)3DCNN (c)SSRN (d)SSFTT (e)MASSFormer (f)SQSFormer (g)SSMamba (h) ours}
    \label{figure4}
\end{figure}
\begin{table}[htbp]
\centering
\footnotesize  
\setlength{\tabcolsep}{3pt}  
\caption{Test data classification results (averages) on the WHU-Hi-HongHu dataset}\label{table3}
\begin{tabular}{@{}l*{8}{c}@{}}
\hline
\multicolumn{1}{c}{Class}  & SVM   & 3DCNN & SSRN  & SSFTT & MASSFormer & SQSFormer & SSMamba & Ours  \\ \hline
1      & 55.32 & 88.12 & 94.57 & 92.20 & 92.26      & 94.77     & \textbf{96.14}   & 95.70 \\
2      & 41.55 & 86.24 & 82.41 & 93.69 & \textbf{96.84}      & 96.40     & 95.76   & 84.43 \\
3      & 35.05 & 86.83 & 91.46 & 88.72 & 86.27      & 90.62     & 91.32   & \textbf{92.96} \\
4      & 33.42 & 89.89 & 96.18 & 96.30 & 94.71      & 97.41     & 98.01   & \textbf{98.24} \\
5      & 50.31 & 91.62 & 96.82 & 93.83 & 94.52      & 95.59     & \textbf{99.30}   & 97.75 \\
6      & 34.78 & 93.75 & 94.54 & 93.64 & 93.61      & 93.96     & 95.04   & \textbf{96.32} \\
7      & 20.54 & 67.57 & 77.54 & 87.12 & 80.45      & 85.14     & 85.59   & \textbf{88.68} \\
8      & 24.42 & 86.20 & 96.84 & 89.55 & 90.07      & 91.29     & \textbf{97.46}   & 94.83 \\
9      & 67.10 & 94.74 & 96.56 & 95.37 & 93.78      & 94.72     & 96.21   & \textbf{97.91} \\
10     & 22.99 & 71.09 & 71.19 & 88.54 & 89.44      & 92.06     & 91.33   & \textbf{92.73} \\
11     & 20.52 & 80.62 & 78.08 & 85.38 & 87.59      & 88.50     & 87.92   & \textbf{93.30} \\
12     & 20.84 & 74.36 & 86.34 & 82.61 & 82.86      & 88.67     & \textbf{89.47}   & 85.44 \\
13     & 20.64 & 73.92 & 74.88 & 76.48 & 79.23      & \textbf{84.40}     & 80.07   & 80.40 \\
14     & 31.76 & 87.28 & 95.43 & 94.51 & 95.33      & \textbf{96.67}      & 93.30   & 87.63 \\
15     & 69.38 & 97.37 & 98.79 & 99.55 & \textbf{99.57}      & 98.48     & 99.49   & 97.63 \\
16     & 52.98 & 92.01 & 94.78 & 96.69 & 94.98      & 96.54     & 97.09   & \textbf{98.81} \\
17     & 53.73 & 89.22 & \textbf{99.97} & 96.61 & 98.66      & 99.31     & 99.95   & 98.22 \\
18     & 42.45 & 98.53 & 98.97 & 98.15 & 98.56      & \textbf{99.50}     & 99.20   & 98.81 \\
19     & 36.38 & 89.98 & 93.58 & 94.10 & 96.09      & \textbf{96.11}     & 94.09   & 93.70 \\
20     & 24.18 & 88.79 & 93.48 & 93.65 & 96.45      & 97.77     & \textbf{99.53}   & 98.96 \\
21     & 44.71 & 98.47 & 99.89 & 97.69 & 99.37      & 99.72     & 99.83   & \textbf{100}   \\
22     & 38.00 & 96.08 & 97.14 & 96.43 & 98.57      & 97.25     & 99.70   & \textbf{99.95} \\  \hline
OA(\%) & 33.84 & 86.80 & 91.58 & 92.59 & 91.72      & 94.19     & 94.56   & \textbf{94.94} \\
AA(\%) & 38.23 & 87.39 & 91.34 & 92.31 & 92.69      & 94.31     & 94.82   & \textbf{94.92} \\
K×100  & 26.76 & 83.73 & 89.46 & 90.71 & 89.65      & 92.70     & 93.15   & \textbf{93.63} \\  \hline
\end{tabular}
\end{table}
\begin{figure}[htbp]
    \centering
    \includegraphics[width=1\linewidth]{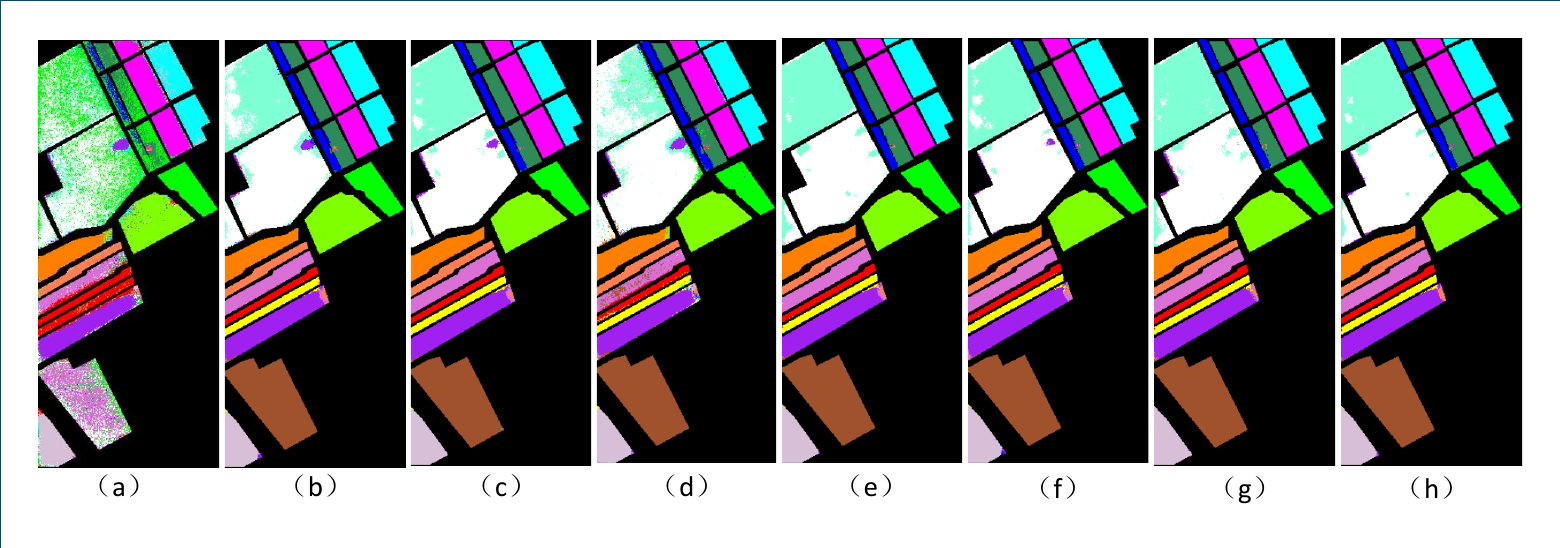}  
    \caption{Classification results of different methods on the Salinas dataset (a)SVM (b)3DCNN (c)SSRN (d)SSFTT (e)MASSFormer (f)SQSFormer (g)SSMamba (h) ours}
    \label{figure5}
\end{figure}
\begin{table}[htbp]
\centering
\footnotesize  
\setlength{\tabcolsep}{3pt}  
\caption{Test data classification results (averages) on the Salinas dataset}\label{table4}
\begin{tabular}{lllllllll}
\hline
\multicolumn{1}{c}{Class}  & SVM   & 3DCNN & SSRN  & SSFTT & MASSFormer & SQSFormer & SSMamba & Ours  \\ \hline
1      & 92.70 & 100   & 96.38 & 94.02 & 100        & 100       & 97.98   & \textbf{100}   \\
2      & 98.61 & 98.53 & 96.56 & 99.23 & 99.90      & 98.02     & 97.99   & \textbf{99.95} \\
3      & 75.17 & 97.38 & 95.55 & 95.26 & 100        & 96.43     & 97.96   & \textbf{100}   \\
4      & 96.79 & 98.12 & 98.72 & 98.73 & 96.87      & 97.99     & 97.83   & \textbf{99.63} \\
5      & 91.04 & 97.13 & 99.59 & 98.20 & \textbf{99.89}      & 97.86     & 96.63   & 98.11 \\
6      & 99.87 & 97.89 & 98.37 & 99.68 & 98.95      & 94.46     & 97.99   & \textbf{99.97} \\
7      & 94.30 & 96.64 & 99.73 & 96.71 & 99.89      & 98.99     & \textbf{100}     & 99.97 \\
8      & 65.65 & 98.32 & \textbf{100}   & 80.18 & 97.90      & 92.46     & 89.43   & 95.12 \\
9      & 95.03 & 95.95 & 97.17 & 98.07 & 100        & 100       & 100     & \textbf{100}   \\
10     & 80.87 & \textbf{100}   & 98.13 & 86.49 & 99.76      & 99.02     & 97.39   & 95.01 \\
11     & 58.82 & 96.13 & 98.14 & 69.47 & 99.70      & 98.78     & 99.78   & \textbf{100}   \\
12     & 86.41 & 97.35 & 99.63 & 97.33 & 99.92      & 100       & 100     & \textbf{100}   \\
13     & 81.66 & 94.74 & 97.85 & 92.40 & 88.34      & \textbf{99.93}     & 99.76   & 99.89 \\
14     & 80.08 & 97.62 & 98.54 & 97.65 & 93.62      & 98.46     & 96.75   & \textbf{99.71} \\
15     & 48.14 & \textbf{98.33} & 94.19 & 66.09 & 91.59      & 96.56     & 94.73   & 95.99 \\
16     & 88.65 & 97.86 & 95.36 & 93.31 & \textbf{99.99}      & 98.65     & 96.88   & 98.99 \\ \hline
OA(\%) & 80.50 & 95.14 & 96.15 & 87.95 & 97.92      & 97.63     & 95.60   & \textbf{98.35} \\
AA(\%) & 83.36 & 96.08 & 97.89 & 91.43 & 97.91      & 97.79     & 96.97   & \textbf{98.99} \\ 
K×100  & 78.21 & 94.03 & 96.05 & 86.62 & 97.68      & 96.54     & 95.34   & \textbf{98.16} \\ \hline
\end{tabular}
\end{table}

It can also be seen intuitively from Figures \ref{figure3}-\ref{figure5} that the classification result graphs on the three datasets show that the performance of SVM is the worst, indicating that its classification performance is the poorest. The methods based on CNN and Transformer are better than SVM, but it's difficult to determine who is better. Among them, the classification effect of SSMamba and our method MSCM-net is the best. The specific results analysis reveals that Table 4 presents the classification results on the Indian Pines dataset. It is evident that the method proposed in this study achieves the highest performance in all metrics. OA reaches 94.35\%. AA is 96.31\%. And the Kappa value is 93.03\%. Compared with the suboptimal method, OA and AA increased by 2.29\% and 0.58\% respectively. On the WHU-Hi-HongHu dataset, as shown in Table 5, the method proposed achieves the highest OA of 94.94\% and Kappa value of 93.63\%, outperforming all other comparison methods. In Table 6, the method presented demonstrates the best classification performance on the Salinas dataset. Compared with all the other methods listed, the OA, AA, and kappa values are all improved by more than 2 percentage points.

Overall, our method achieves the best results in all three datasets as evaluated by the OA, AA and Kappa metrics. For specific categories, it also achieves the majority of the optimal accuracy, which is sufficient to demonstrate that our method has certain competitiveness and performance advantages.
\subsection{Parameter analysis}
\begin{figure}[htbp]
    \centering
    \includegraphics[width=1\linewidth]{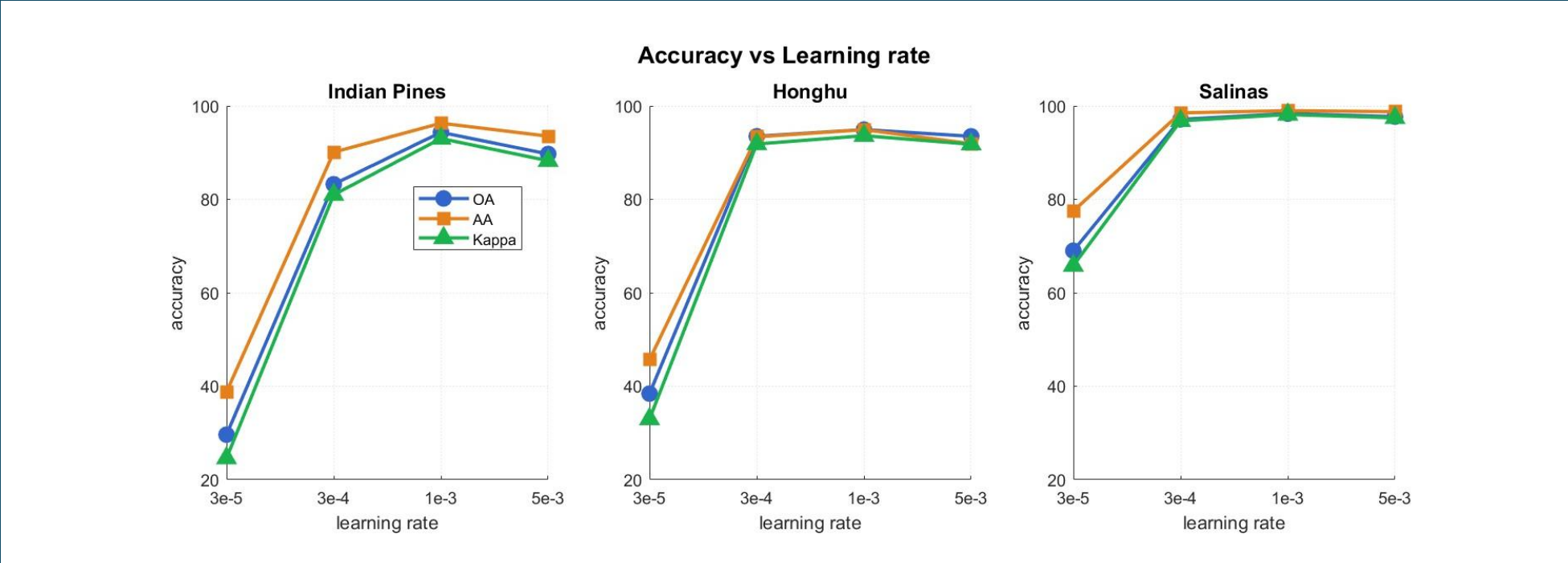}  
    \caption{Line graph showing the impact of learning rate on accuracy on three datasets}
    \label{figure6}
\end{figure}

\begin{figure}[htbp]
    \centering
    \includegraphics[width=1\linewidth]{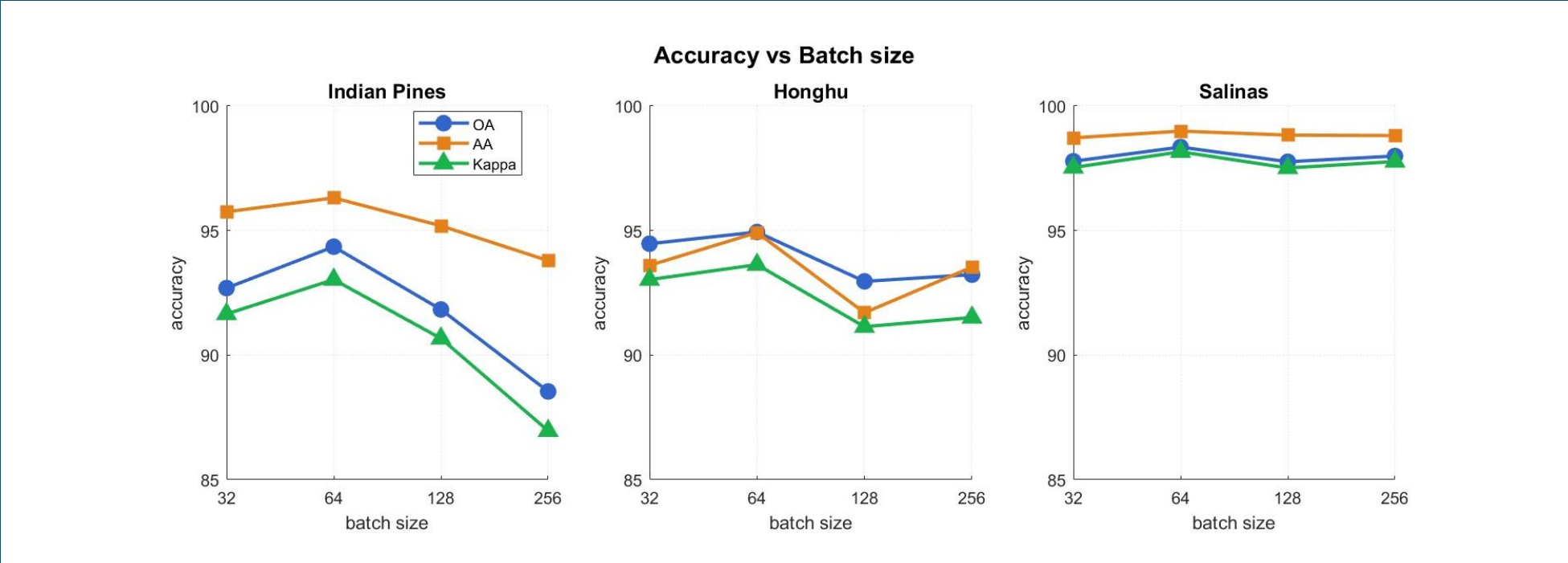}  
    \caption{Line graph showing the impact of batch size on accuracy across three datasets}
    \label{figure7}
\end{figure}

\begin{figure}[htbp]
    \centering
    \includegraphics[width=1\linewidth]{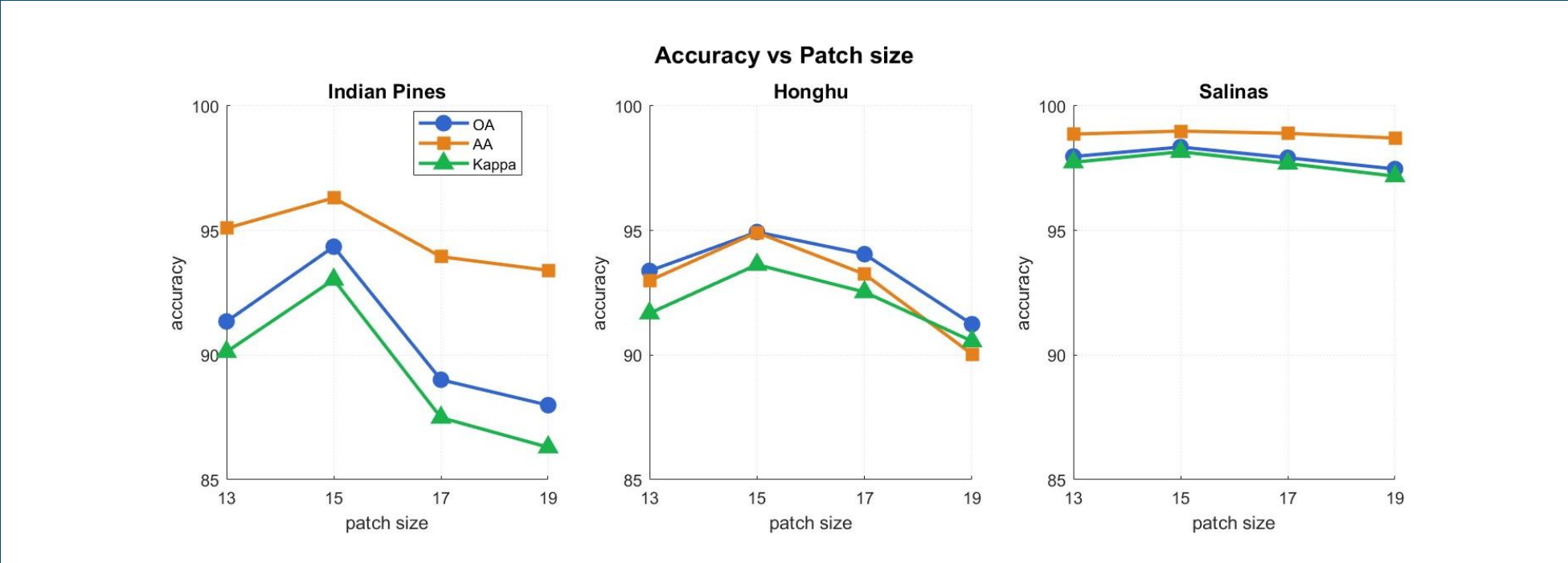}  
    \caption{Line graph showing the influence of patch size on accuracy across the three datasets}
    \label{figure8}
\end{figure}

\begin{figure}[htbp]
    \centering
    \includegraphics[width=1\linewidth]{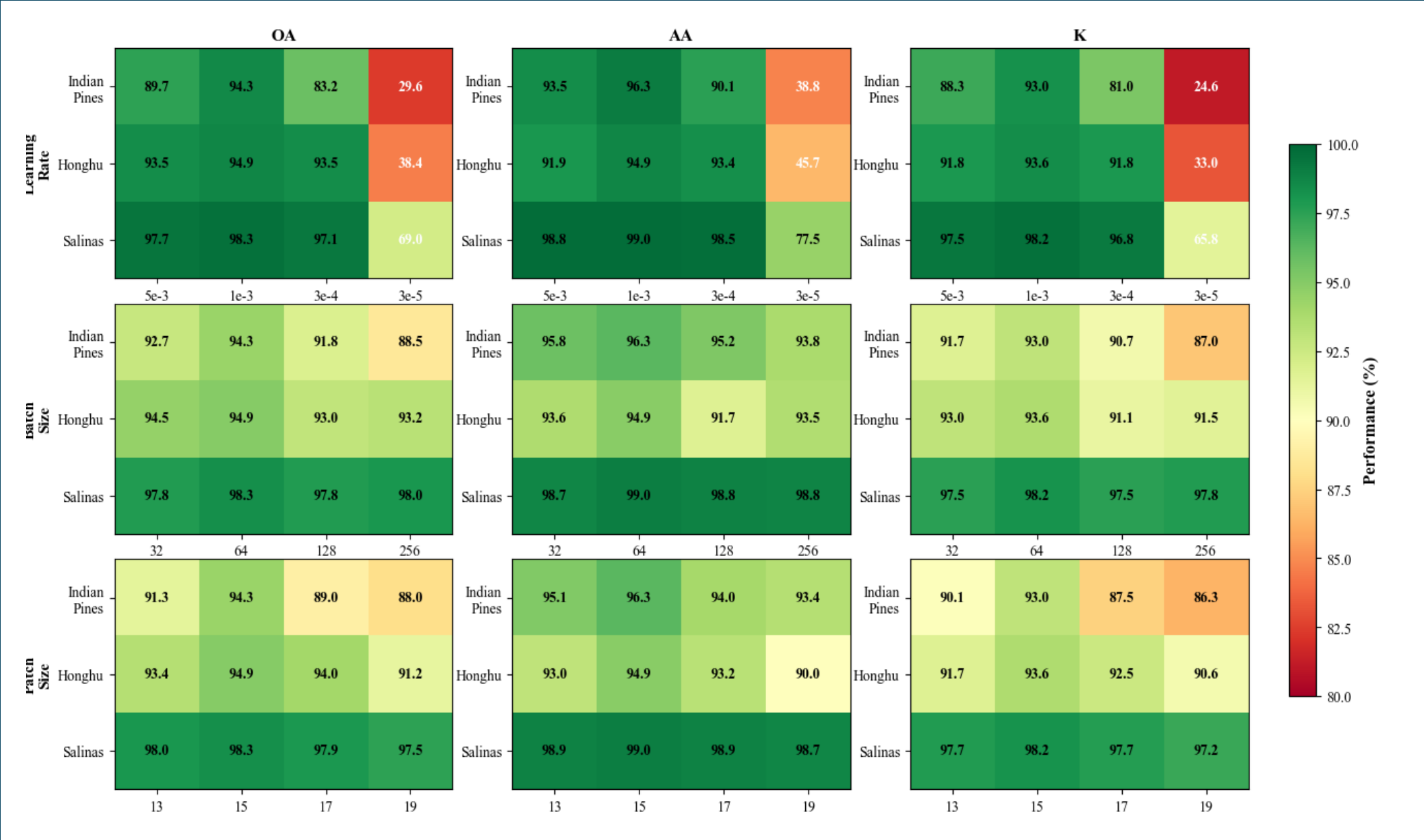}  
    \caption{Heatmap showing the impact of hyperparameters on the classification performance of the three datasets}
    \label{figure9}
\end{figure}

In this section, we analyze the influence of several key hyperparameters of the MSCM-net model on the model's performance, namely learning rate, batch size, and patch size. To facilitate the analysis, we plot line graphs as shown in Figure \ref{figure6}-\ref{figure8} and the heatmap as shown in Figure \ref{figure9}.

The results shown in Figure \ref{figure6} indicate that initially, as the learning rate is gradually increased from a small value of 3e-5, OA of the model also gradually improves. When the learning rate is set to 0.001, the model's performance reaches its peak. Further increasing the learning rate does not help in improving the model's accuracy. Therefore, the optimal learning rate chosen in this article is 0.001.

The results shown in Figure \ref{figure7} indicate that as the batch size increases, the model initially benefits from richer representations, thereby improving accuracy. However, beyond a certain point, further increasing the batch size leads to a decline in performance. This decline may be due to the model beginning to capture noise rather than useful information, resulting in a decrease in generalization performance. In our model, when the batch size is set to 64, the model accuracy reaches the highest on all three datasets. This value achieves a balance between underfitting and overfitting for the model, providing the best result.

The results shown in Figure \ref{figure8} indicate that, across all three datasets, the impact of patch size on the model performance follows a similar pattern. In Figure 10, initially, as the patch size increases, the performance of the model significantly improves. However, after reaching a certain point, the performance begins to stagnate or even slightly decline. This indicates that there is an optimal range of patch sizes within which the model can capture sufficient contextual information without introducing unnecessary complexity or noise. Through experiments, we finally choose the value 15, which performs the best on all three datasets, as the optimal patch size.

Furthermore, to conduct a more comprehensive analysis of the model's sensitivity to key hyperparameters, we have plotted a $3\times 3$ heat map matrix, as shown in Figure \ref{figure9}. Each row of the heat map corresponds to a specific hyperparameter, and each column represents a performance metric. The color intensity of each cell represents the classification performance. Warm colors (green, yellow) indicate higher accuracy, while cool colors (red) indicate lower performance. As shown in Figure \ref{figure9}, there is a clearly defined "optimal range" for performance at a learning rate of 1e-3. An excessively high learning rate (5e-3) leads to unstable training, while an excessively low learning rate (3e-5) results in insufficient convergence. When the batch size is set to 64, the best results are achieved. A smaller batch size (32) introduces higher variance, while a larger batch size (128, 256) reduces the generalization ability of the model. The optimal patch size is $15\times 15$. A smaller patch size ($13\times 13$) captures insufficient contextual information, while a larger patch size ($17\times 17$, $19\times 19$) introduces excessive background noise, slightly reducing the classification accuracy.
\subsection{Performance and complexity analysis(on the Indian Pines dataset)}
The parameter quantity and model complexity of all methods are analyzed on the Indian Pines dataset, and the analysis results are recorded in Table \ref{table5}. From the table, it can be seen that compared with all the comparison methods, the MSCM-net model proposed has the smallest number of parameters, the shortest inference time, and the highest OA. This clearly indicates that our model has achieved a good balance between parameter efficiency and model performance.

The MSCM-net has only 200,000 parameters, which is much lighter than other Transformer-based models (such as SQSFormer with 1,045,000 parameters and CNN-based models such as SSRN with 292,000 parameters). Our method not only has a lower complexity but also achieves the highest OA of 94.35\%, outperforming these more complex models. This efficiency highlights the ability of the model in this paper to capture basic features without relying on a large parameter space.

Furthermore, MSCM-net requires only 0.19 GFLOPs. Although it is not the lowest, it is significantly lower than most other models, such as SSMamba (2.844G) and 3DCNN (1.091G). Although SSFTT has the lowest FLOPs, it fails to achieve the accuracy of the model in this study. This indicates that the model in this paper effectively balances the computational cost and performance.

Regarding the inference speed, the method based on Mamba has a faster inference speed, such as the method we present and SSMamba. This is because Mamba does not have the problem of quadratic complexity, as its computing speed is faster than that of models based on Transformers. Moreover, Mamba also employs a hardware-aware algorithm, fully utilizing the GPU of the computer, which further accelerates the inference speed.

In conclusion, MSCM-net demonstrates its ability to achieve high accuracy through a more streamlined and computationally efficient design. By achieving better performance with fewer parameters and lower computational requirements, MSCM-net is particularly suitable for practical applications with limited computing resources.
\begin{table}[htbp]
\centering
\footnotesize  
\setlength{\tabcolsep}{3pt}  
\caption{Analysis of Model Parameters and Complexity of Different Methods on the Indian Pines Dataset}\label{table5}
\small  
\begin{tabular}{@{}l|cccccccc@{}}
\hline
\multicolumn{1}{c}{}   & SVM & 3DCNN & SSRN & SSFTT & MASSFormer & SQSFormer & SSMamba & Ours \\ \hline
Parameters (M) & / & 0.564 & 0.292 & 0.915 & 0.304 & 1.045 & 0.407 & 0.201 \\
FLOPs (G) & / & 1.091 & 0.246 & 0.078 & 0.817 & 0.233 & 2.844 & 0.19 \\
Test time (s) & / & 5.99 & 6.54 & 1.48 & 5.203 & 12.024 & 2.11 & 1.2 \\
OA (\%) & 69.34 & 89.26 & 91.70 & 90.18 & 92.06 & 93.19 & 91.59 & 94.35 \\  \hline
\end{tabular}
\end{table}

\section{Ablation experiments}
In this section, we conduct ablation experiments on the three key modules used in the model of this paper to verify their effectiveness and their contribution to the overall framework.

(1) The number of Mambas

\begin{figure}[htbp]
    \centering
    \includegraphics[width=1\linewidth]{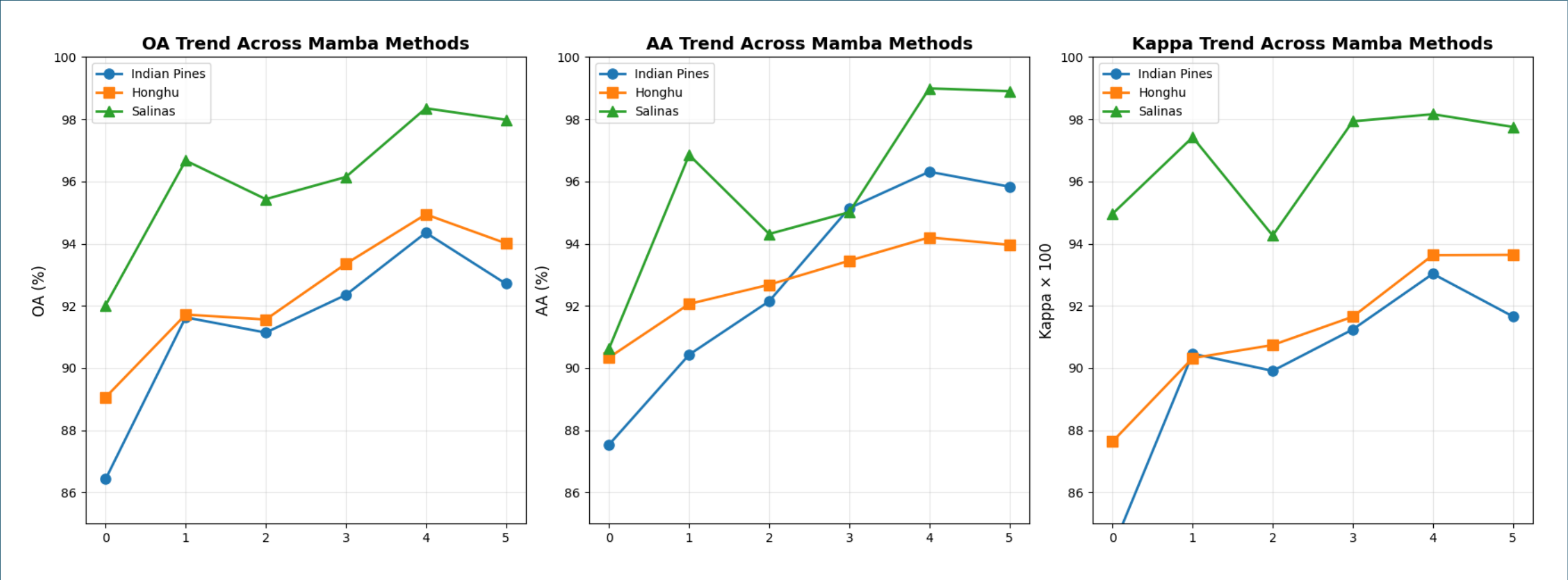}  
    \caption{Line graph showing the ablation experiment results of Mamba block quantity}
    \label{figure10}
\end{figure}

To investigate the impact of the depth of the Mamba block on the model performance, we conduct ablation experiments on the Mamba module, setting the number of Mamba blocks from 0 to 5. Experiments are conducted on three datasets, and the experimental results are plotted as a line graph as shown in Figure \ref{figure10}. As shown in Figure \ref{figure10}, when the number of Mamba blocks ranges from 0 to 1, on all three datasets, the slope of the line is the greatest, showing a clear upward trend. This indicates that the values of OA, AA and kappa in the model without using the Mamba block are significantly lower than those with the Mamba block, suggesting that the Mamba block added to the model in this study has greatly contributed to improving the model's accuracy. Furthermore, it can be observed that as the number of Mamba blocks increases, the classification accuracy shows a trend of first rising and then falling. When the number of Mamba blocks is 4, it achieves the best performance on all three datasets. However, when the number of Mamba blocks is increased to 5, there is no significant improvement in accuracy. Therefore, considering the complexity of the model, it is ultimately decided that the number of Mamba blocks in this model is 4.

(2) SeNet

To further enhance the feature extraction capability of the main module MCSE in our method, the SeNet is incorporated into the MCSE module. And the effectiveness of SeNet is verified through experiments. Table \ref{table6}'s ablation experiments demonstrate the classification results on three datasets with and without SeNet in the MCSE module. As can be seen from Table 8, on the Indian Pines dataset, compared with not incorporating SeNet, after adding SeNet, OA, AA and Kappa have increased by 4.29, 2.36 and 4.35 percentage points respectively. On the Salinas dataset, the OA, AA and Kappa metrics increase by 2.51, 0.83 and 2.79 percentage points respectively. This is sufficient to prove the role that the SeNet module plays in the MCSE.
\begin{table}[htbp]
\centering
\caption{Ablation Experiments of SENet}
    \label{table6}
\begin{tabular}{l|l|lll}
\hline
    &        & Indian Pines & Honghu & Salinas \\ \hline
    & OA(\%) & 94.35        & 94.94  & 98.35   \\
Yes & AA(\%) & 96.31        & 94.20  & 98.99   \\
    & K×100  & 93.03        & 93.63  & 98.16   \\ \hline
    & OA(\%) & 90.06        & 93.73  & 95.84   \\
No  & AA(\%) & 93.95        & 93.57  & 98.16   \\
    & K×100  & 88.68        & 92.12  & 95.37   \\ \hline
\end{tabular}
\end{table}

(3) residual connection

To stabilize the input of the Mamba block, residual connections are added to the model, specifically before the input Mamba block. This arrangement ensures that the input of the Mamba block not only contains the features processed by the MCSE module but also the original features. The purpose is to guarantee that the input contains sufficient information for Mamba to make a choice. To verify the effectiveness of this residual connection, we conducted corresponding ablation experiments for this improvement. The experimental results are shown in Table \ref{table7}. Analysis of the experimental data reveals that when residual connections were not used, the OA, AA and Kappa values on the three datasets were all lower than the results obtained when residual connections were employed. This indicates that adding residual connections at this position in the model provides the Mamba module with more comprehensive input features. It not only retains the spectral specificity of hyperspectral data but also incorporates rich contextual information, laying a solid foundation for the subsequent deep sequential modeling of Mamba, thereby effectively improving the overall classification performance of the model.
\begin{table}[htbp]
\centering
\caption{Abandonment Experiment of Residual Connection}
    \label{table7}
\begin{tabular}{l|l|lll}
\hline
    &        & Indian Pines & Honghu & Salinas \\ \hline
    & OA(\%) & 94.35        & 94.94  & 98.35   \\
Yes & AA(\%) & 96.31        & 94.20  & 98.99   \\
    & K×100  & 93.03        & 93.63  & 98.16   \\ \hline
    & OA(\%) & 92.18        & 93.56  & 97.75   \\
No  & AA(\%) & 95.36        & 92.98  & 98.52   \\
    & K×100  & 92.87        & 92.64  & 97.04   \\ \hline
\end{tabular}
\end{table}

(4) Dual-branch Feature Aggregation Module  

In order to verify the impact of the dual-branch center and global feature aggregation module proposed in the model on the improvement of model accuracy, detailed ablation experiments are conducted on this module on three datasets. The specific approach is as follows: separately add only the central feature extraction branch to the model, only the global feature extraction branch to the model, and a dual branch of central feature extraction and global feature extraction. OA on the three datasets is recorded respectively. The experimental results are shown in Table \ref{table8}. It is clearly observed that the overall accuracy of using both the central extraction and global extraction branches is higher than that of using only one of the branches alone. Among them, the effect of using only the center extraction is the poorest. This is because the center extraction only contains the key information and lacks the completeness of the data. In contrast, the global feature extraction can pay more attention to the global information that can identify the categories, thereby enabling classification. Therefore, the optimal choice is to combine the center extraction and the global extraction, which is also the approach adopted in this paper.
\begin{table}[htbp]
\centering
\caption{Abandonment Experiment of the Dual-Branch Feature Aggregation Module}
    \label{table8}
\begin{tabular}{ll|lll}
\hline
center   extraction & global   extraction & Indian   Pines & HongHu & Salinas \\ \hline
√                   &                     & 86.36          & 85.22  & 94.65   \\
                    & √                   & 93.42          & 93.86  & 97.29   \\
√                   & √                   & 94.35          & 94.94  & 98.35   \\ \hline
\end{tabular}
\end{table}

\section{Conclusions}
We have proposed MSCM-net, a hybrid framework integrating multi-scale convolutional neural networks with selective state-space models for hyperspectral image classification. By combining multi-scale convolution's spatial feature extraction advantages with Mamba's capacity to capture remote spectral dependencies, MSCM-net effectively processes both local and global information in hyperspectral data. The framework employs multi-scale convolutional modules with different kernel sizes to extract multi-range features, enabling more accurate hyperspectral data representation. Multiple stacked Mamba blocks perform sequence modeling to extract deeper features, while residual connections prevent original information loss. A dual-branch feature aggregation structure with center and global extraction enhances the model's ability to capture both central pixel and edge information. Extensive experiments on benchmark datasets demonstrate that MSCM-net achieves higher classification accuracy with fewer parameters compared to state-of-the-art methods, balancing powerful feature extraction with computational efficiency. This study validates the potential of combining selective state-space models with multi-scale CNNs for hyperspectral image classification, offering a valuable research direction for future work in this field.











\bibliographystyle{elsarticle-num-names} 
\bibliography{cas-refs}

@article{ GaoJingpeng2026,
Author = {Gao, Jingpeng and Chen, Geng and Ji, Xiangyu and Shen, Chen},
Title = {Reinforced graph aggregation cross-domain few-shot learning for
   hyperspectral remote sensing image classification},
Journal = {SIGNAL PROCESSING},
Year = {2026},
Volume = {238},
Month = {JAN},
DOI = {10.1016/j.sigpro.2025.110101},
EarlyAccessDate = {JUN 2025},
Article-Number = {110101},
ISSN = {0165-1684},
EISSN = {1872-7557},
ORCID-Numbers = {Ji, Xiangyu/0009-0007-0910-4307
   Chen, Geng/0009-0004-7989-279X},
Unique-ID = {WOS:001506342400002},
}

@article{ ZhaoJinling2025,
Author = {Zhao, Jinling and Wu, Keke and Zhang, Lu and Huang, Wenjiang and Ruan,
   Chao and Huang, Linsheng},
Title = {Patch-based hierarchical residual spectral-spatial convolutional network
   for hyperspectral image classification},
Journal = {SIGNAL PROCESSING},
Year = {2025},
Volume = {230},
Month = {MAY},
DOI = {10.1016/j.sigpro.2024.109850},
EarlyAccessDate = {DEC 2024},
Article-Number = {109850},
ISSN = {0165-1684},
EISSN = {1872-7557},
ResearcherID-Numbers = {huang, lin/OIU-2389-2025},
ORCID-Numbers = {Zhao, Jinling/0000-0002-8352-7689
   },
Unique-ID = {WOS:001391675300001},
}

@article{ WangXinya2023,
Author = {Wang, Xinya and Hu, Qian and Cheng, Yingsong and Ma, Jiayi},
Title = {Hyperspectral Image Super-Resolution Meets Deep Learning: A Survey and
   Perspective},
Journal = {IEEE-CAA JOURNAL OF AUTOMATICA SINICA},
Year = {2023},
Volume = {10},
Number = {8},
Pages = {1668-1691},
Month = {AUG},
DOI = {10.1109/JAS.2023.123681},
ISSN = {2329-9266},
EISSN = {2329-9274},
ResearcherID-Numbers = {Wang, Xinya/KIA-3984-2024
   Ma, Jiayi/Y-2470-2019},
ORCID-Numbers = {Hu, Qian/0009-0005-7613-988X
   Wang, Xinya/0000-0003-2144-9811
   cheng, yingsong/0009-0002-8054-4886
   Ma, Jiayi/0000-0003-3264-3265},
Unique-ID = {WOS:001037849500003},
}

@article{ Yang2024,
Author = {Yang, Xu and Ma, Kejia and Zhang, Dejia and Song, Shaozhong and An,
   Xiaofeng},
Title = {Classification of soybean seeds based on RGB reconstruction of
   hyperspectral images},
Journal = {PLOS ONE},
Year = {2024},
Volume = {19},
Number = {9},
Month = {SEP 4},
DOI = {10.1371/journal.pone.0307329},
Article-Number = {e0307329},
EISSN = {1932-6203},
Unique-ID = {WOS:001308434600043},
}

@article{ Ali2024,
Author = {Ali, Faizan and Razzaq, Ali and Tariq, Waheed and Hameed, Akhtar and
   Rehman, Abdul and Razzaq, Khizar and Sarfraz, Sohaib and Rajput, Nasir
   Ahmed and Zaki, Haitham E. M. and Shahid, Muhammad Shafiq and Ondrasek,
   Gabrijel},
Title = {Spectral Intelligence: AI-Driven Hyperspectral Imaging for Agricultural
   and Ecosystem Applications},
Journal = {AGRONOMY-BASEL},
Year = {2024},
Volume = {14},
Number = {10},
Month = {OCT},
DOI = {10.3390/agronomy14102260},
Article-Number = {2260},
EISSN = {2073-4395},
ResearcherID-Numbers = {Hameed, Akhtar/ABG-5082-2021
   Zaki, Haitham/S-5056-2018
   Onasek, Gabrijel/V-3625-2019
   Shahid, Muhammad Shafiq/AAI-2253-2020
   Rajput, Nasir Ahmed/ABI-4090-2020
   Razzaq, Khizar/AAA-8668-2021
   Razzaq, Ali/JUV-2734-2023},
ORCID-Numbers = {Hameed, Akhtar/0000-0003-0434-2457
   Zaki, Haitham/0000-0001-7245-830X
   Onasek, Gabrijel/0000-0001-8398-0099
   Shahid, Muhammad Shafiq/0000-0002-3550-0000
   Rajput, Nasir Ahmed/0000-0003-3507-2605
   },
Unique-ID = {WOS:001342429300001},
}

@article{ WangJinfeng2023,
Author = {Wang, Jinfeng and Chen, Guoqing and Ju, Jinyan and Lin, Tenghui and
   Wang, Ruidong and Wang, Zhentao},
Title = {Characterization and classification of urban weed species in northeast
   China using terrestrial hyperspectral images},
Journal = {WEED SCIENCE},
Year = {2023},
Volume = {71},
Number = {4},
Pages = {353-368},
Month = {JUL},
DOI = {10.1017/wsc.2023.36},
EarlyAccessDate = {AUG 2023},
Article-Number = {PII S004317452300036X},
ISSN = {0043-1745},
EISSN = {1550-2759},
ORCID-Numbers = {Wang, Zhentao/0000-0002-0613-9527
   Chen, Guoqing/0009-0007-3821-2649},
Unique-ID = {WOS:001041628300001},
}

@article{ Song2023,
Author = {Song Ruo-xi and Feng Yi-ning and Cheng Wei and Wang Xiang-hai},
Title = {Advance in Hyperspectral Images Change Detection},
Journal = {SPECTROSCOPY AND SPECTRAL ANALYSIS},
Year = {2023},
Volume = {43},
Number = {8},
Pages = {2354-2362},
Month = {AUG},
DOI = {10.3964/j.issn.1000-0593(2023)08-2354-09},
ISSN = {1000-0593},
ResearcherID-Numbers = {Wang, Xianghai/GRR-4512-2022},
Unique-ID = {WOS:001054445900005},
}

@article{ DuPeijun2012,
Author = {Du, Peijun and Tan, Kun and Xing, Xiaoshi},
Title = {A novel binary tree support vector machine for hyperspectral remote
   sensing image classification},
Journal = {OPTICS COMMUNICATIONS},
Year = {2012},
Volume = {285},
Number = {13-14},
Pages = {3054-3060},
Month = {JUN 15},
DOI = {10.1016/j.optcom.2012.02.092},
ISSN = {0030-4018},
EISSN = {1873-0310},
ORCID-Numbers = {Du, Peijun/0000-0002-2488-2656
   Tan, Kun/0000-0001-6353-0146},
Unique-ID = {WOS:000303629900015},
}

@article{ ZhangYouqiang2018,
Author = {Zhang, Youqiang and Cao, Guo and Li, Xuesong and Wang, Bisheng},
Title = {Cascaded Random Forest for Hyperspectral Image Classification},
Journal = {IEEE JOURNAL OF SELECTED TOPICS IN APPLIED EARTH OBSERVATIONS AND REMOTE
   SENSING},
Year = {2018},
Volume = {11},
Number = {4, SI},
Pages = {1082-1094},
Month = {APR},
DOI = {10.1109/JSTARS.2018.2809781},
ISSN = {1939-1404},
EISSN = {2151-1535},
ResearcherID-Numbers = {Zhang, Youqiang/ISV-0771-2023
   Cao, Guo/AAC-1388-2022},
ORCID-Numbers = {Zhang, Youqiang/0000-0002-4761-4726
   Cao, Guo/0000-0002-2689-0932},
Unique-ID = {WOS:000429956000007},
}

@article{ LiuGuangxin2022,
Author = {Liu, Guangxin and Wang, Liguo and Liu, Danfeng and Fei, Lei and Yang,
   Jinghui},
Title = {Hyperspectral Image Classification Based on Non-Parallel Support Vector
   Machine},
Journal = {REMOTE SENSING},
Year = {2022},
Volume = {14},
Number = {10},
Month = {MAY},
DOI = {10.3390/rs14102447},
Article-Number = {2447},
EISSN = {2072-4292},
Unique-ID = {WOS:000802721000001},
}

@article{ ZhouDing-Xuan2018,
Author = {Zhou, Ding-Xuan},
Title = {Deep distributed convolutional neural networks: Universality},
Journal = {ANALYSIS AND APPLICATIONS},
Year = {2018},
Volume = {16},
Number = {6},
Pages = {895-919},
Month = {NOV},
DOI = {10.1142/S0219530518500124},
ISSN = {0219-5305},
EISSN = {1793-6861},
ResearcherID-Numbers = {Zhou, Ding-Xuan/B-3160-2013},
Unique-ID = {WOS:000449149300005},
}

@article{ YasinMagombe2024,
Author = {Yasin, Magombe and Sarigul, Mehmet and Avci, Mutlu},
Title = {Logarithmic Learning Differential Convolutional Neural Network},
Journal = {NEURAL NETWORKS},
Year = {2024},
Volume = {172},
Month = {APR},
DOI = {10.1016/j.neunet.2024.106114},
EarlyAccessDate = {JAN 2024},
Article-Number = {106114},
ISSN = {0893-6080},
EISSN = {1879-2782},
ResearcherID-Numbers = {AVCI, MUTLU/F-9898-2018
   Sarıgül, Mehmet/ISA-7474-2023},
ORCID-Numbers = {Yasin, Magombe/0000-0001-8200-4440
   AVCI, MUTLU/0000-0002-4412-4764
   Sarıgül, Mehmet/0000-0001-7323-6864},
Unique-ID = {WOS:001163206400001},
}

@article{ AthaDeegan2018,
Author = {Atha, Deegan J. and Jahanshahi, Mohammad R.},
Title = {Evaluation of deep learning approaches based on convolutional neural
   networks for corrosion detection},
Journal = {STRUCTURAL HEALTH MONITORING-AN INTERNATIONAL JOURNAL},
Year = {2018},
Volume = {17},
Number = {5},
Pages = {1110-1128},
Month = {SEP},
DOI = {10.1177/1475921717737051},
ISSN = {1475-9217},
EISSN = {1741-3168},
Unique-ID = {WOS:000443737800006},
}

@article{ YangXiaofei2018,
Author = {Yang, Xiaofei and Ye, Yunming and Li, Xutao and Lau, Raymond Y. K. and
   Zhang, Xiaofeng and Huang, Xiaohui},
Title = {Hyperspectral Image Classification With Deep Learning Models},
Journal = {IEEE TRANSACTIONS ON GEOSCIENCE AND REMOTE SENSING},
Year = {2018},
Volume = {56},
Number = {9},
Pages = {5408-5423},
Month = {SEP},
DOI = {10.1109/TGRS.2018.2815613},
ISSN = {0196-2892},
EISSN = {1558-0644},
ResearcherID-Numbers = {huang, xiaohui/KRP-2903-2024
   Ye, Yunming/N-8557-2015
   Yang, Xiaofei/AAA-8220-2021},
ORCID-Numbers = {Lau, Raymond/0000-0002-5751-4550
   Zhang, Xiaofeng/0000-0003-0972-8842
   Yang, Xiaofei/0000-0003-2458-6774},
Unique-ID = {WOS:000443147600035},
}

@article{ PaulArati2021,
Author = {Paul, Arati and Bhoumik, Sanghamita and Chaki, Nabendu},
Title = {SSNET: an improved deep hybrid network for hyperspectral image
   classification},
Journal = {NEURAL COMPUTING \& APPLICATIONS},
Year = {2021},
Volume = {33},
Number = {5},
Pages = {1575-1585},
Month = {MAR},
DOI = {10.1007/s00521-020-05069-1},
EarlyAccessDate = {JUN 2020},
ISSN = {0941-0643},
EISSN = {1433-3058},
ResearcherID-Numbers = {CHAKI, NABENDU/A-5869-2015},
Unique-ID = {WOS:000541031000002},
}

@article{ FengFan2019,
Author = {Feng, Fan and Wang, Shuangting and Wang, Chunyang and Zhang, Jin},
Title = {Learning Deep Hierarchical Spatial-Spectral Features for Hyperspectral
   Image Classification Based on Residual 3D-2D CNN},
Journal = {SENSORS},
Year = {2019},
Volume = {19},
Number = {23},
Month = {DEC},
DOI = {10.3390/s19235276},
Article-Number = {5276},
EISSN = {1424-8220},
ORCID-Numbers = {Feng, Fan/0000-0003-1344-3727},
Unique-ID = {WOS:000507606200219},
}

@article{ LeiXiaohan2025,
Author = {Lei, Xiaohan and Xie, Fuding and Jin, Cui},
Title = {Gram Angle Field-Based Siamese Graph Convolutional Neural Network for
   Hyperspectral Images Classification},
Journal = {IEEE GEOSCIENCE AND REMOTE SENSING LETTERS},
Year = {2025},
Volume = {22},
DOI = {10.1109/LGRS.2024.3521461},
Article-Number = {5501305},
ISSN = {1545-598X},
EISSN = {1558-0571},
ORCID-Numbers = {Xie, Fuding/0000-0003-1392-2699
   Lei, Xiaohan/0009-0005-5193-1694},
Unique-ID = {WOS:001398604700003},
}

@article{ HuXiang2021,
Author = {Hu, Xiang and Li, Teng and Zhou, Tong and Liu, Yu and Peng, Yuanxi},
Title = {Contrastive Learning Based on Transformer for Hyperspectral Image
   Classification},
Journal = {APPLIED SCIENCES-BASEL},
Year = {2021},
Volume = {11},
Number = {18},
Month = {SEP},
DOI = {10.3390/app11188670},
Article-Number = {8670},
EISSN = {2076-3417},
ORCID-Numbers = {Hu, Xiang/0000-0002-1798-8508},
Unique-ID = {WOS:000699188200001},
}

@article{ TanYunfei2024,
Author = {Tan, Yunfei and Li, Ming and Yuan, Longfa and Shi, Chaoshan and Luo,
   Yonghang and Wen, Guihao},
Title = {Hyperspectral image classification with embedded linear vision
   transformer},
Journal = {EARTH SCIENCE INFORMATICS},
Year = {2024},
Volume = {18},
Number = {1},
Month = {DEC 18},
DOI = {10.1007/s12145-024-01651-6},
Article-Number = {69},
ISSN = {1865-0473},
EISSN = {1865-0481},
ResearcherID-Numbers = {Luo, Yonghang/KEZ-8458-2024
   YUAN, LONGFA/NVM-0939-2025},
Unique-ID = {WOS:001380076200001},
}

@article{ FuChuan2025,
Author = {Fu, Chuan and Zhou, Tianyuan and Guo, Tan and Zhu, Qikui and Luo, Fulin
   and Du, Bo},
Title = {CNN-Transformer and Channel-Spatial Attention based network for
   hyperspectral image classification with few samples},
Journal = {NEURAL NETWORKS},
Year = {2025},
Volume = {186},
Month = {JUN},
DOI = {10.1016/j.neunet.2025.107283},
EarlyAccessDate = {FEB 2025},
Article-Number = {107283},
ISSN = {0893-6080},
EISSN = {1879-2782},
ResearcherID-Numbers = {Du, Bo/AEY-9731-2022
   Zhou, Tianyuan/JRZ-1315-2023
   Luo, Fulin/Y-8972-2019
   Zhu, Qikui/HHN-1681-2022},
ORCID-Numbers = {Du, Bo/0000-0001-8104-3448
   Zhou, Tianyuan/0009-0007-7083-4288
   },
Unique-ID = {WOS:001435063000001},
}

@article{ FanXiao-yong2025,
Author = {Fan, Xiao-yong and Li, Heng-kai and Liu, Kun-ming and Wang, Xin-li and
   Yu, Yang and Li, Xino-yu},
Title = {A Multi-Layer Attention Convolutional Neural Network Model for Fine
   Classification of Hyperspectral Images in Rare Earth Mining Areas},
Journal = {SPECTROSCOPY AND SPECTRAL ANALYSIS},
Year = {2025},
Volume = {45},
Number = {9},
Pages = {2666-2675},
Month = {SEP},
DOI = {10.3964/j.issn.1000-0593(2025)09-2666-10},
ISSN = {1000-0593},
ResearcherID-Numbers = {hengkai, li/HKO-5418-2023},
Unique-ID = {WOS:001633545300035},
}

@article{ Schuessler2022,
Author = {Schuessler, Max},
Title = {Machine learning with nonlinear state space models},
Journal = {AT-AUTOMATISIERUNGSTECHNIK},
Year = {2022},
Volume = {70},
Number = {11},
Pages = {1027-1028},
Month = {NOV 25},
DOI = {10.1515/auto-2022-0089},
ISSN = {0178-2312},
EISSN = {2196-677X},
Unique-ID = {WOS:000883806300010},
}

@article{ SunJunding2025,
Author = {Sun, Junding and Chen, Kaixin and Wang, Shuihua and Zhang, Yudong and
   Xu, Zhaozhao and Wu, Xiaosheng and Tang, Chaosheng},
Title = {DGFE-Mamba: Mamba-Based 2D Image Segmentation Network},
Journal = {JOURNAL OF BIONIC ENGINEERING},
Year = {2025},
Volume = {22},
Number = {4},
Pages = {2135-2150},
Month = {JUL},
DOI = {10.1007/s42235-025-00711-x},
EarlyAccessDate = {APR 2025},
ISSN = {1672-6529},
EISSN = {2543-2141},
ResearcherID-Numbers = {Zhang, Yudong/I-7633-2013
   Tang, Chao-Sheng/E-8498-2019},
Unique-ID = {WOS:001476168600001},
}

@article{ ZhangChongbin2025,
Author = {Zhang, Chongbin and Zheng, Jiaxiang and Cao, Moxi},
Title = {A music source separation method integrating time-frequency decoupling
   and mamba-based state space modeling},
Journal = {SCIENTIFIC REPORTS},
Year = {2025},
Volume = {15},
Number = {1},
Month = {OCT 16},
DOI = {10.1038/s41598-025-20179-3},
Article-Number = {36280},
ISSN = {2045-2322},
ORCID-Numbers = {ZHENG, JIAXIANG/0009-0000-1407-8894},
Unique-ID = {WOS:001596677400035},
}

@article{ WangHao2025,
Author = {Wang, Hao and Zhuang, Peixian and Zhang, Xiaochen and Li, Jiangyun},
Title = {DBMGNet: A Dual-Branch Mamba-GCN Network for Hyperspectral Image
   Classification},
Journal = {IEEE TRANSACTIONS ON GEOSCIENCE AND REMOTE SENSING},
Year = {2025},
Volume = {63},
DOI = {10.1109/TGRS.2025.3564364},
Article-Number = {4410517},
ISSN = {0196-2892},
EISSN = {1558-0644},
ResearcherID-Numbers = {Zhang, Xiaochen/HHS-9663-2022},
ORCID-Numbers = {Wang, Hao/0009-0002-8865-8589
   Zhang, Xiaochen/0009-0002-5905-8541},
Unique-ID = {WOS:001498239100007},
}

@article{ JiangYonghua2025,
Author = {Jiang, Yonghua and Zhang, Shuai and Wang, Chengjun and Zhang, Guo and
   Tan, Meilin and Du, Bin and Shen, Xin},
Title = {SEDGM: A Structure-Enhanced Spatial-Spectral Dynamic Gating Mamba for
   Hyperspectral Image Classification},
Journal = {IEEE TRANSACTIONS ON GEOSCIENCE AND REMOTE SENSING},
Year = {2025},
Volume = {63},
DOI = {10.1109/TGRS.2025.3626930},
Article-Number = {5532422},
ISSN = {0196-2892},
EISSN = {1558-0644},
ResearcherID-Numbers = {Wang, Chengjun/W-7920-2019},
ORCID-Numbers = {Shen, Xin/0000-0002-9692-822X
   ZHANG, GUO/0000-0002-3987-5336
   Zhang, Shuai/0009-0009-8579-8640
   },
Unique-ID = {WOS:001616590700001},
}

@article{ LiYapeng2024,
Author = {Li, Yapeng and Luo, Yong and Zhang, Lefei and Wang, Zengmao and Du, Bo},
Title = {MambaHSI: Spatial-Spectral Mamba for Hyperspectral Image Classification},
Journal = {IEEE TRANSACTIONS ON GEOSCIENCE AND REMOTE SENSING},
Year = {2024},
Volume = {62},
DOI = {10.1109/TGRS.2024.3430985},
Article-Number = {5524216},
ISSN = {0196-2892},
EISSN = {1558-0644},
ResearcherID-Numbers = {Li, Yapeng/KWU-8292-2024
   Du, Bo/AEY-9731-2022
   Zhang, Lefei/HHM-8850-2022
   Luo, Yong/G-7969-2016},
ORCID-Numbers = {Li, Yapeng/0009-0000-2983-885X
   Du, Bo/0000-0001-8104-3448
   wang, zengmao/0000-0002-9326-0316
   Luo, Yong/0000-0002-2296-6370},
Unique-ID = {WOS:001285022800027},
}

@article{ LiangLianhui2025,
Author = {Liang, Lianhui and Xie, Peiyi and Zhang, Ying and Li, Jiaxin and Zhang,
   Zhe and Li, Jun and Plaza, Antonio},
Title = {DBMLLA: Double-Branch Mamba-Like Linear Attention Network for
   Hyperspectral Image Classification},
Journal = {IEEE TRANSACTIONS ON GEOSCIENCE AND REMOTE SENSING},
Year = {2025},
Volume = {63},
DOI = {10.1109/TGRS.2025.3605798},
Article-Number = {5524315},
ISSN = {0196-2892},
EISSN = {1558-0644},
ResearcherID-Numbers = {李(微信 BatAug), 嘉鑫/KQU-1808-2024
   Plaza, Antonio/C-4455-2008
   Zhang, Ying/JHU-4844-2023},
ORCID-Numbers = {Zhang, Ying/0000-0002-3972-3316
   Xie, Peiyi/0009-0004-0329-7089
   Liang, Lianhui/0000-0001-6958-0443
   李, 嘉鑫/0000-0002-1237-542X
   Plaza, Antonio/0000-0002-9613-1659
   },
Unique-ID = {WOS:001575188400050},
}

@article{ KongFanqiang2024,
Author = {Kong, Fanqiang and Huang, Murong and Tang, Jiahui and Ren, Guanglong},
Title = {End-to-end feature domain residual coding network for multispectral
   image compression based on interspectral prediction},
Journal = {JOURNAL OF APPLIED REMOTE SENSING},
Year = {2024},
Volume = {18},
Number = {3},
Month = {JUL 1},
DOI = {10.1117/1.JRS.18.036508},
EISSN = {1931-3195},
Unique-ID = {WOS:001330379800003},
}

@article{ WuJiann-Ming2008,
Author = {Wu, Jiann-Ming},
Title = {Multilayer Potts Perceptrons With Levenberg-Marquardt Learning},
Journal = {IEEE TRANSACTIONS ON NEURAL NETWORKS},
Year = {2008},
Volume = {19},
Number = {12},
Pages = {2032-2043},
Month = {DEC},
DOI = {10.1109/TNN.2008.2003271},
ISSN = {1045-9227},
EISSN = {1941-0093},
Unique-ID = {WOS:000261544900004},
}

@article{ TanKun2008,
Author = {Tan Kun and Du Pei-Jun},
Title = {Hyperspectral remote sensing image classification based on support
   vector machine},
Journal = {JOURNAL OF INFRARED AND MILLIMETER WAVES},
Year = {2008},
Volume = {27},
Number = {2},
Pages = {123-128},
Month = {APR},
DOI = {10.3724/SP.J.1010.2008.00123},
ISSN = {1001-9014},
ResearcherID-Numbers = {Tan, Kun/AAA-6254-2019},
ORCID-Numbers = {Du, Peijun/0000-0002-2488-2656
   Tan, Kun/0000-0001-6353-0146},
Unique-ID = {WOS:000255675800010},
}

@article{ FiratHuseyin2023,
Author = {Firat, Huseyin and Hanbay, Davut},
Title = {Comparison of 3D CNN based deep learning architectures using
   hyperspectral images},
Journal = {JOURNAL OF THE FACULTY OF ENGINEERING AND ARCHITECTURE OF GAZI
   UNIVERSITY},
Year = {2023},
Volume = {38},
Number = {1},
Pages = {521-534},
DOI = {10.17341/gazimmfd.977688},
ISSN = {1300-1884},
EISSN = {1304-4915},
ResearcherID-Numbers = {Hanbay, Davut/AAG-8511-2019
   FIRAT, Hüseyin/ABB-7417-2021},
ORCID-Numbers = {Hanbay, Davut/0000-0003-2271-7865
   FIRAT, Hüseyin/0000-0002-1257-8518},
Unique-ID = {WOS:000835332900041},
}

@article{ ZhongZilong2018,
Author = {Zhong, Zilong and Li, Jonathan and Luo, Zhiming and Chapman, Michael},
Title = {Spectral-Spatial Residual Network for Hyperspectral Image
   Classification: A 3-D Deep Learning Framework},
Journal = {IEEE TRANSACTIONS ON GEOSCIENCE AND REMOTE SENSING},
Year = {2018},
Volume = {56},
Number = {2},
Pages = {847-858},
Month = {FEB},
DOI = {10.1109/TGRS.2017.2755542},
ISSN = {0196-2892},
EISSN = {1558-0644},
ResearcherID-Numbers = {LI, Jonathan/AAA-7712-2021},
ORCID-Numbers = {Zhong, Zilong/0000-0003-0104-9116
   Luo, Zhiming/0000-0002-3411-9582
   LI, Jonathan/0000-0001-7899-0049},
Unique-ID = {WOS:000424627500019},
}

@article{ SunLe2022,
Author = {Sun, Le and Zhao, Guangrui and Zheng, Yuhui and Wu, Zebin},
Title = {SpectralSpatial Feature Tokenization Transformer for Hyperspectral Image
   Classification},
Journal = {IEEE TRANSACTIONS ON GEOSCIENCE AND REMOTE SENSING},
Year = {2022},
Volume = {60},
DOI = {10.1109/TGRS.2022.3144158},
Article-Number = {5522214},
ISSN = {0196-2892},
EISSN = {1558-0644},
ResearcherID-Numbers = {Zheng, Yuhui/AAF-2420-2019},
ORCID-Numbers = {Guang Rui, Zhao/0000-0001-9613-9645
   Sun, Le/0000-0001-6465-8678
   Wu, Zebin/0000-0002-7162-0202
   },
Unique-ID = {WOS:000770073000026},
}

@article{ SunLe2024,
Author = {Sun, Le and Zhang, Hang and Zheng, Yuhui and Wu, Zebin and Ye, Zhonglin
   and Zhao, Haixing},
Title = {MASSFormer: Memory-Augmented Spectral-Spatial Transformer for
   Hyperspectral Image Classification},
Journal = {IEEE TRANSACTIONS ON GEOSCIENCE AND REMOTE SENSING},
Year = {2024},
Volume = {62},
DOI = {10.1109/TGRS.2024.3392264},
Article-Number = {5516415},
ISSN = {0196-2892},
EISSN = {1558-0644},
ResearcherID-Numbers = {Zheng, Yuhui/AAF-2420-2019},
ORCID-Numbers = {Wu, Zebin/0000-0002-7162-0202
   Sun, Le/0000-0001-6465-8678
   },
Unique-ID = {WOS:001217067100019},
}

@article{ ChenNing2024,
Author = {Chen, Ning and Fang, Leyuan and Xia, Yang and Xia, Shaobo and Liu, Hui
   and Yue, Jun},
Title = {Spectral Query Spatial: Revisiting the Role of Center Pixel in
   Transformer for Hyperspectral Image Classification},
Journal = {IEEE TRANSACTIONS ON GEOSCIENCE AND REMOTE SENSING},
Year = {2024},
Volume = {62},
DOI = {10.1109/TGRS.2024.3361652},
Article-Number = {5402714},
ISSN = {0196-2892},
EISSN = {1558-0644},
ResearcherID-Numbers = {Fang, Leyuan/G-1468-2011
   Yue, Jun/AFG-8947-2022
   Xia, Yang/NXX-5395-2025
   },
ORCID-Numbers = {Fang, Leyuan/0000-0003-2351-4461
   Yue, Jun/0000-0002-6465-5052
   Xia, Yang/0009-0000-8742-9157
   chen, ning/0009-0001-9280-975X},
Unique-ID = {WOS:001173263900045},
}

@article{ HuangLingbo2024,
Author = {Huang, Lingbo and Chen, Yushi and He, Xin},
Title = {Spectral-Spatial Mamba for Hyperspectral Image Classification},
Journal = {REMOTE SENSING},
Year = {2024},
Volume = {16},
Number = {13},
Month = {JUL},
DOI = {10.3390/rs16132449},
Article-Number = {2449},
EISSN = {2072-4292},
ResearcherID-Numbers = {Chen, Yushi/ACI-9252-2022
   },
ORCID-Numbers = {Chen, Yushi/0000-0003-2421-0996
   He, Xin/0000-0003-0455-4230},
Unique-ID = {WOS:001269994900001},
}
\section*{Declaration of competing interest}
The authors declare that they have no known competing financial interests or personal relationships that could have appeared to
influence the work reported in this paper.
\section*{Acknowledgments}
This work was supported by the Natural Science Foundation of
Shandong Province (ZR2021QF113), the Outstanding Youth Innovation
Team in Shandong Higher Education Institutions (2022KJ162).



\end{document}